\pdfoutput=1

\documentclass[11pt]{article}
\usepackage[final]{acl}

\usepackage{times}
\usepackage{latexsym}
\usepackage{graphicx}
\usepackage{multicol}
\usepackage{multirow}
\usepackage{adjustbox}
\usepackage{comment}
\usepackage{caption}
\usepackage{mathtools}
\usepackage{amsmath}
\usepackage{algorithm}
\usepackage{float}
\usepackage{url}
\usepackage[justification=centering]{caption}
\usepackage{algpseudocode}
\algblock{Input}{EndInput}
\algnotext{EndInput}
\algblock{Paramters}{EndParameter}
\algnotext{EndParameter}

\usepackage[T1]{fontenc}

\usepackage[utf8]{inputenc}

\usepackage{microtype}

\usepackage{inconsolata}
\newcommand{\noteng}[1]{\textcolor{magenta}{\textbf{NG:} #1}}
\newcommand{\pg}[1]{\textcolor{red}{\textbf{PG:} #1}}
\newcommand{\ankan}[1]{\textcolor{violet}{\textbf{ankan:} #1}}

\newcommand{\sombit}[1]{\textcolor{green}{\textbf{sombit:} #1}}

\title{Introducing Spotlight: A Novel Approach for Generating Captivating Key Information from Documents}

\author{$^{1*}$Ankan Mullick \qquad $^{1*}$Sombit Bose \qquad $^1$Rounak Saha \qquad $^2$Ayan Kumar Bhowmick \qquad \\{\bf $^2$Aditya Vempaty \hspace{3mm} $^2$Prasenjit Dey \hspace{3mm} $^2$Ravi Kokku \hspace{3mm} $^1$Pawan Goyal \hspace{3mm} $^1$ Niloy Ganguly} \\ \texttt{\{ankanm, sbcs.sombit.24, runk20\}@kgpian.iitkgp.ac.in}\\ \texttt{\{pawang, niloy\}@cse.iitkgp.ac.in}, \texttt{\{ayan, aditya, prasenjit, ravi\}@emergence.ai}\\ $^1$IIT Kharagpur, India.  $^2$Emergence AI}

\begin{document}


\maketitle
\def\thefootnote{*}\footnotetext{Equal Contribution. }
\begin{abstract}
Analyzing and processing vast amounts of textual data presents significant challenges in efficiently extracting key information.
In this paper, we introduce \textbf{`Spotlight'}, a novel paradigm for information extraction that produces concise, engaging narratives by highlighting the most compelling aspects of a document. Unlike highlights (fragmented key points) and traditional summaries, which prioritize comprehensive coverage, spotlights selectively emphasize intriguing content to foster deeper reader engagement with the source material. We formally differentiate spotlights from related constructs and support our analysis with a detailed benchmarking study using new datasets curated for this work. To generate high-quality spotlights, we propose a two-stage approach: fine-tuning a large language model on our benchmark data, followed by alignment via Direct Preference Optimization (DPO). Our comprehensive evaluation demonstrates that the resulting model not only identifies key elements with precision but also enhances readability and boosts the engagement value of the original document. Datasets and code are available at \url{https://github.com/ankan2/Spotlight-EMNLP2025}.
\end{abstract}

\begin{table*}[htb]
\centering
\captionsetup{justification=raggedright,singlelinecheck=off}
\begin{adjustbox}{width=0.95\linewidth}    
\begin{tabular}{c|c|c|c|c|c|c||c|c|c}
\hline
   & \textbf{ Info-Dist} & \textbf{Compactness} & \textbf{Readability}  & \textbf{Extractive} & \textbf{Informative} & \textbf{Interest-Generation} & \textbf{Faithful}   & \textbf{Length} &  \textbf{Mini-Story}\\ \hline
Summary$\dagger$  & Uniform & High & Medium & Low &  High & Low & High & Long  & Yes\\ \hline
Teaser & Non-uniform & Low & High & Low &  Low & High & Low   & Short  & No \\ \hline
Headline$\mathsection$ & Non-uniform & Low & High & High & Low & Medium & Medium   & Short  & No \\ \hline
\begin{tabular}{@{}l@{}}  Highlight    \end{tabular}    & Non-uniform & Low & High & High & High & Medium & High   & Medium &  No\\ \hline
Spotlight  & Skewed & Low & High & High & High &  High & High   & Long  & Yes\\\hline
\end{tabular}
\vspace{-3mm}
\end{adjustbox}
\caption{Qualitative Analysis of different metrics to evaluate and characterize various methods (Summary, Teaser, Headline, Highlight, and Spotlight) for deriving information from a source document. \newline  
\textbf{Information Distribution (Info-Dist)}:   Uniformity with which information is extracted from different sections of the document; 
\textbf{Compactness}: Details in Section \ref{compact}B; ~~\textbf{Readbility}: Details in Section \ref{compact}C; 
\newline 
\textbf{Extractive}: The degree to which the generated content retains phrases or segments from the original text, rather than paraphrasing or reinterpreting them.; 
 \newline
 \textbf{Informative}: Extent to which the generated content contains information relevant to the original text;\newline
 \textbf{Interest-Generation}: Capturing the reader’s attention and/or stimulating curiosity to encourage engagement with the source document; \newline 
 \textbf{Faithful}: Degree to which the content is exclusively derived from the source document; \newline
\textbf{Length}: Total number of words, sentences, or characters it contains; \newline \textbf{Mini-Story}: A brief compact narrative that encapsulates the main plot or theme of an article.  \newline$\dagger$: We consider the case of abstractive summary. $\mathsection$: For non-clickbaitish Headlines.}
\label{tab:spotlight-vs-others}
\end{table*}

\section{Introduction}

In the field of document processing, information condensation techniques such as summarization aim to extract and present the essential content of lengthy texts in a more accessible and concise format. A summary is typically a structured and comprehensive abstraction that encapsulates all major points, providing readers with a faithful overview of the document’s primary content. In contrast, this paper introduces spotlight—a novel, self-contained narrative designed to highlight the most engaging and insightful elements of a document. Unlike traditional summaries that prioritize full coverage, a spotlight constructs a coherent and compelling mini-narrative that remains faithful to the original text, with the goal of capturing the reader’s interest and encouraging further exploration of the complete document.

Spotlights differ from traditional forms of document processing such as highlights, headlines, and teasers, which serve specific roles in information extraction. Highlights consist of short, disconnected snippets that reflect key points \cite{arumae2019towards,cho2020better,wei2018utilizing,woodsend2010automatic}, but often lack coherence and narrative flow. Headlines are single-sentence summaries that capture the central theme \cite{colmenares2015heads,ao2021pens,gavrilov2019self}, while teasers are crafted to entice readers by offering intriguing yet vague glimpses of the content \cite{karn2019news,wozny2023generating,chakraborty2016stop,deng2024prompt}, without revealing substantive information.

This paper introduces the concept of spotlights as an innovative step forward from traditional summaries and highlights. A spotlight is a concise, self-contained narrative, typically prepared by the document’s author, that highlights its most significant and engaging aspects.
Unlike traditional formats, spotlights offer the narrative depth of summaries while uniquely aiming to generate interest in the full text.\footnote{Note that even the summarization can be target-specific, some examples include query focused summarization~\cite{vig2022exploring,baumel2016topic,liu2024querysum}, layman summarization~\cite{goldsack2023overview,salaun2022conditional,luo2022readability}, personal-based summarization~\cite{mullick2024persona}, goal oriented~\cite{goldstein2007genre,ham2020end,jin2024comprehensive}, aspect-based summarization~\cite{mullick2024leveraging} etc. summaries with controllable readability~\cite{he2020ctrlsum,zhang2023macsum,ribeiro2023generating}, summarizing documents with respect to key elements~\cite{ryu2024key,wang2023element}. However, the objective of these methods is generally to achieve maximum coverage and deliver a quick, faithful read—potentially sparing the reader from having to engage with the full main text, which appears to be in contrast to the objective of spotlight. Moreover, the concept of spotlights can also be extended to cater to these objectives, e.g., query-focused spotlight, persona-based spotlight etc.} They strike a careful balance between readability, engagement, and fidelity, functioning as standalone narratives. Much like an author's spotlight talk at a conference, they are concise yet compelling, designed to draw readers in and encourage deeper exploration.
While these five condensation techniques may exist along a continuum, they possess distinct characteristics that justify separate classification. Table \ref{tab:spotlight-vs-others} outlines their comparative features, such as source fidelity, level of intrigue, abstractiveness, conciseness, narrative coherence, readability, and the distribution of content coverage.

While extensive research exists on summarization and highlight extraction (see Appendix \ref{rel_works}), spotlight generation remains unexplored. 
In this work, we introduce the concept, quantitatively differentiate spotlights from summaries, and propose effective generation methods. Although no dedicated spotlight datasets currently exist, we identify and adapt existing sources—such as news articles, CSPubSum, Wikipedia, and conference presentations—to align with spotlight characteristics. These are compared with corresponding similar concepts (summary, teaser, headline, highlight based on the framework defined through Table~\ref{tab:spotlight-vs-others}.



We further develop a cost-effective, domain-specific spotlight generation method using a fine-tuned Large Language Model with Direct Preference Optimization (DPO). Our empirical evaluation, which includes benchmark datasets and AI-based critique mechanisms, shows that our approach outperforms both prompt-based techniques and larger models such as those from the GPT family across several key metrics.


The major contributions of this paper are as follows:
(a) We introduce the concept of spotlight generation and formally define the problem;
(b) We compile four datasets and benchmark spotlight vis-a-vis summary. 
(c) We propose a preference-based model for spotlight generation and perform extensive experiments to establish its effectiveness.

\section{Dataset}

We curate four datasets — News Category Dataset, CSPubSum, Wikipedia, and Research Presentation. These datasets were selected because each includes an additional condensed abstraction that highlights the most compelling aspects of the document, aligning well with the objectives of spotlight generation. Additionally, to highlight the distinction between a spotlight and a summary, we also created summaries for the corresponding documents.

\subsection{Spotlight Datasets}
\noindent \textbf{News Category Dataset~\cite{misra2022news}:} This dataset has fine-grained category information for large number of news articles (published between 2012 and 2022) and we have randomly selected 14,080 news headlines for our experiment. 
We consider the headline along with a short description provided as a spotlight of the news article as shown in Fig. \ref{fig:news_cat_spot} (Appendix).
The rationale is that headlines and the associated short descriptions provide a concise, informative structure in an engaging manner, similar to the desired spotlights. 

\noindent 

\noindent \textbf{CSPubSum~\cite{collins2017supervised}:} This dataset comprises the abstract, introduction, and conclusion sections of various research articles, accompanied by author-curated key points that highlight the core contributions of each paper. These author-curated key points are used as spotlights in our framework. 


\noindent \textbf{Wikipedia:}  
We curate Wikipedia articles across four popular domains - Healthcare, Music, Education, and Life \& Career—organized 
as detailed in Table~\ref{tab:domain_categories_set} (Appendix). 
Each article is divided into category-specific sections, with the introductory paragraphs emphasizing the most compelling aspects of each section.  
To assign a portion of this paragraph to a specific category, we extract the sentence(s) most similar to the category’s focus. Thus we have each section as a separate document and the corresponding extracted sentences as its spotlight. The total number of documents obtained 
is 17,323.

\noindent \textbf{Research Presentation:} We curate this dataset by  collecting 1,230 audio transcripts of NeurIPS 2022 \footnote{\url{https://nips.cc/virtual/2022/events/Spotlight}} and 2023 \footnote{\url{https://nips.cc/virtual/2023/events/spotlight-posters-2023}} spotlight Poster presentations. These audio transcripts are derived from 5-minute self-narrative videos, which focus on presenting the most intriguing and important aspects of the research papers. This concise format is meant to capture the attention of the audience and encourage them to explore the full paper. Since these audio transcripts are generated from automated transcription services, they often contain errors that could hinder readability and interpretation. Therefore, we apply transcript cleaning techniques like contraction expansion, punctuation removal, etc. to improve the quality and utility of these transcripts. 

\subsection{Summary Generation} 
We generate summaries manually by annotators as well as by using LLMs. 

\noindent {\bf Human Generated Summaries:}
For deeper analysis, first, we manually create a smaller set of summaries for four datasets. A total of 60 articles were randomly selected from each dataset and assigned to four annotators (one annotator per dataset). Each annotator independently generated summaries for the assigned articles, adhering to the average length of the spotlight sections of the original articles. Detailed steps are in Appendix \ref{prolific_summ_gen}. 

\noindent {\bf LLM Generated Summaries:}
Since manual summary generation is time-consuming and costly, to scale up the summary generation task, we utilize GPT-4 to automate the creation of summaries for all articles of entire datasets involved in the experiments. The GPT-4-generated summaries had an average length of approximately 15\% of the original articles, closely aligning with the average spotlight length of 13.76\% of the original articles.
To ensure the quality of these summaries, we implemented a three-step automated validation process (Table \ref{tab:filtering})  followed by manual verification. Approximately 3.64\% of the document-summary pairs were filtered out based on Table \ref{tab:filtering}.

Additionally, we evaluate the similarity between the manually created summaries and those generated by GPT-4 for the 60 articles that were curated through both methods. The results demonstrate a high degree of alignment, with 
BERT-F1 scores of 81.5\%, and GPT-4-based similarity scores of 89.8\%. Thus, it shows that GPT-4 generated summaries are 
close to human-curated summaries and can be used as a replacement. 

\begin{table}[ht]
\centering
\begin{adjustbox}{width=\linewidth}
\begin{tabular}{|l|c|}
\hline
\textbf{Filtering Steps with Criteria (removed)} & \textbf{\%} \\ \hline
Step 1: Too many special characters and other string (HTML tags and \# )	&  1.45 \\ \hline
Step 2: Incomplete Summary (By checking punctuations)  & 1.07  \\ \hline
\begin{tabular}{@{}l@{}} Step 3: If the summary contains Terms or numbers not present in the document 
\\ \end{tabular} &  1.12\\\hline
Overall summaries filtered out	&  3.64  \\ \hline
\end{tabular}
\end{adjustbox}
\vspace{-2mm}
\caption{Step-by-Step Data Filtering}
\vspace{-3mm}
\label{tab:filtering}
\end{table}

\section{Spotlight: Characterization} \label{spotlight-define}

As can be seen in Table \ref{tab:spotlight-vs-others}, the concept of spotlight differs from the summary on the first six features, while it is similar on the last three features. We thus study the differences between the first six and then characterize the last three features. 

\begin{table*}[!ht]
\vspace{-2mm}
\centering
\begin{adjustbox}{width=\linewidth}
\begin{tabular}{|c|c|cc|c|ccc|c|c|c|c|c|}
\hline
\multicolumn{2}{|c|}{Dataset} & \multicolumn{2}{c|}{A. Distribution}                                                                               & B. Compact- & \multicolumn{3}{c|}{C. Readability (\%)}   &  D. Informati- & E. Degree of  & F. Interest-   & G. Factuality(\%) & H. Length\\\cline{3-4}\cline{6-8}
                  \multicolumn{2}{|c|}{} & \multicolumn{1}{c|}{Entropy$\downarrow$} & \multicolumn{1}{c|}{Skew50$\uparrow$} &  ness$\downarrow$ & \multicolumn{1}{c|}{Flesch-Kincaid$\downarrow$} & \multicolumn{1}{c|}{Flesch Reading$\uparrow$} & \multicolumn{1}{c|}{Gunning$\downarrow$ } & veness          & Extraction  & Generation (\%) $\uparrow$    & &  \\\hline \hline
              \multirow{2}{*}{News} & Spot  & \multicolumn{1}{c|}{0.69}                  & \multicolumn{1}{c|}{0.61}                            &      0.52            & \multicolumn{1}{c|}{10.57}                  & \multicolumn{1}{c|}{15.46}                  & \multicolumn{1}{c|}{16.81}                   & 0.061 & 0.021     &   80  &  91.4  & 48.3  \\ \cline{2-13}
               &  Summ    & \multicolumn{1}{c|}{0.96}                  & \multicolumn{1}{c|}{0.02}                              &          1.53         & \multicolumn{1}{c|}{16.89}                  & \multicolumn{1}{c|}{9.45}                  & \multicolumn{1}{c|}{20.10}                               & 0.143 & 0.011     &     24   &   87.2  & 60.1 \\ \hline \hline
                \multirow{2}{*}{CSPubSum} & Spot  & \multicolumn{1}{c|}{0.62}                  & \multicolumn{1}{c|}{0.47}                                   &        0.62           & \multicolumn{1}{c|}{16.28}                  & \multicolumn{1}{c|}{9.67}                  & \multicolumn{1}{c|}{19.73}                    & 0.042 &  0.109     &        86 &   85.3 & 39.3 \\ \cline{2-13}
                & Summ   & \multicolumn{1}{c|}{0.96}                  & \multicolumn{1}{c|}{0.01}                                  &         1.28          & \multicolumn{1}{c|}{20.72}                  & \multicolumn{1}{c|}{6.42}                  & \multicolumn{1}{c|}{23.55}      & 0.112 & 0.100    &      30    &   87.1  & 48.2\\ \hline\hline
              \multirow{2}{*}{Wikipedia}  & Spot    & \multicolumn{1}{c|}{0.76}                  & \multicolumn{1}{c|}{0.29}                        &      0.82             & \multicolumn{1}{c|}{12.62}                  & \multicolumn{1}{c|}{16.26}                  & \multicolumn{1}{c|}{16.37}                      & 0.030 & 0.032  &     86    &  88.2   & 30.1 \\ \cline{2-13}
                & Summ   & \multicolumn{1}{c|}{0.95}                  & \multicolumn{1}{c|}{0.05}                                 &        1.48           & \multicolumn{1}{c|}{16.24}                  & \multicolumn{1}{c|}{9.35}                  & \multicolumn{1}{c|}{19.19}                    &   0.239 & 0.025   &      26     & 84.3 & 43.9 \\ \hline \hline

              Research & Spot   & \multicolumn{1}{c|}{0.65}                  & \multicolumn{1}{c|}{0.36}                        &      0.28             & \multicolumn{1}{c|}{11.84}                  & \multicolumn{1}{c|}{44.10}                  & \multicolumn{1}{c|}{11.85}                     & 0.069 & 0.052     &      82  & 75.3  & 705.2 \\ \cline{2-13}
                Presentation & Summ   & \multicolumn{1}{c|}{0.92}                  & \multicolumn{1}{c|}{0.04}                                 &        0.69           & \multicolumn{1}{c|}{16.03}                  & \multicolumn{1}{c|}{23.83}                  & \multicolumn{1}{c|}{17.21}                    & 0.315 & 0.026   &     24  &  79.3 & 675.9   \\ \hline
\end{tabular}
\end{adjustbox}
\caption{Feature level analysis of 
Spotlight (Spot) vs. Summary (Summ), Entropy, Compactness, Flesch Kincaid and Gunning score lower values signify better, whereas for Skew50, Flesch Reading score and Interest-Generation higher values signify a better result as discussed in Section \ref{spotlight-define}, Additional Results in Appendix Table \ref{tab:qualitative-spot-sum-human} and \ref{tab:degree-infor-extract-human}}
\label{tab:qualitative-spot-sum}
\vspace{-3mm}
\end{table*}

\subsection{Spotlight-Summary:  Distinct Features}
\label{spot-summ-dist-feat}
We do a comparative study on four datasets for 60 annotated samples (each instance has a document, corresponding spotlight and manual summary). 

\noindent \textbf{(A) Information Distribution over Quartiles of the document:} Information distribution of a condensation (summary or spotlight) refers to the 
distribution of key points and essential details of the original document. First, documents are divided into four quadrants and we measure the distribution  
of spotlight or summary information over the four quadrants. We report the average entropy of this distribution and the fraction of documents where 50\% of key points covered by the corresponding spotlight/summary are present in a single quadrant of the document (denote this fraction by Skew50) 
as shown in Table \ref{tab:qualitative-spot-sum}. Our results reveal that spotlights exhibit significantly lower entropy and higher skew compared to summaries, as they tend to focus primarily on a specific (and presumably intriguing) portion of the document.

\noindent \textbf{(B) Information Compactness:} 
\label{compact}
Information compactness refers to the degree to which a condensation distills essential details of the original content. 
We use the average IDF\footnote{IDF stands for Inverse Document Frequency. IDF is a standard metric to measure the informativeness of a word. } score to measure compactness. We calculate the IDF scores for each word of the condensation and then do the average. 
From Table \ref{tab:qualitative-spot-sum} it is clearly seen that the average IDF value is always higher for summary than spotlight indicating more information overload (compactness) in summary than spotlight.

\noindent \textbf{(C) Text Readability:} Text readability is best defined as the ease with which a text can be read and understood in terms of the linguistic features found within a text.  
We utilize six different readability scores, detailed in Table \ref{tab:appendix_readability} (Appendix).

Table \ref{tab:qualitative-spot-sum} shows results of three different readable metrics (due to space limitation, all other details 
are in the Appendix \ref{app:read}) which signifies spotlight has better readability than summary. 
The superior performance of spotlights can be linked to our previous findings. Spotlights convey less information per sentence, which reduces cognitive load and enhances readability. In contrast, summaries tend to be more information-dense, potentially leading to higher cognitive effort for the reader. Next, we examine whether the higher compactness arises from summaries packing more information from the source document.

\noindent \textbf{(D) Informativeness:} 
To quantify the information retained from the source document in a condensation, we extract key-facts from the original documents using key-fact alignment \cite{song-etal-2024-finesure} approach and compare them to those present in spotlights and summaries using exact matching as shown in Table \ref{tab:qualitative-spot-sum} column D. It is seen that the key-facts retained from the document in summary is more than spotlight, which indicates a higher informativeness in summary.


\noindent \textbf{(E) Degree of Extraction:}
To quantify the degree of text reuse in condensations (summaries vs. spotlights), we measure the fraction of sentences lifted near-verbatim from the source document. Specifically, we define a sentence 
$y$ in the condensation as an almost-lift of a document sentence  $x$ if their ROUGE-F1 
score exceeds 0.8 following \cite{bhandari-etal-2020-evaluating}.
Our analysis reveals that spotlights exhibit significantly higher extractiveness than summaries, with a larger fraction of sentences qualifying as almost-lifts (see Table \ref{tab:qualitative-spot-sum}, Column E). This trend holds consistently across ROUGE metrics (R1, R2, RL), as detailed in Appendix Table \ref{tab:degree-infor-extract-human}. We hypothesize that the increased extractiveness of spotlights contributes to their enhanced readability, as they preserve the original document’s phrasing for key compelling content.

\begin{figure*}[!ht]
    \centering
    \includegraphics[width=.99\textwidth]{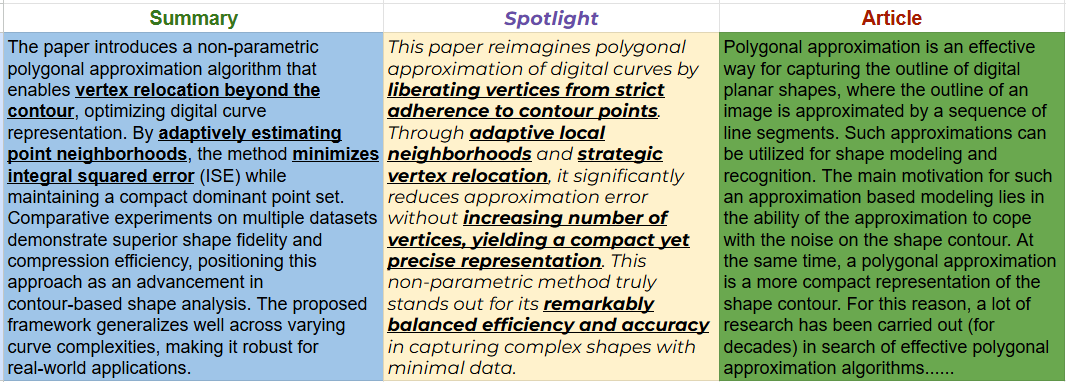}
    \caption{Comparison between a summary and a spotlight generated from the same document. The \textit{summary} presents a concise, technical overview of the algorithm, detailing how adaptive vertex relocation minimizes approximation error with rigorous experimental validation. The \textit{spotlight} describes a captivating mini-story, emphasizing the innovative breakthrough of liberating vertices from the contour to create elegant, engaging shape approximations. It minimizes emphasis on technical methodology and reduces the use of complex terminology, while effectively conveying the significance of the work in a way that encourages readers to explore the main article. Each phrase introduces a novel concept or method, prompting readers to wonder how exactly it works and why it improves results,  thereby triggering a sense of deeper exploration by raising implicit questions in the reader's mind.} The ``interest generation phrases'' are highlighted in \textbf{bold} and \underline{underlined}). 
     \vspace{-3mm}
    \label{fig:spot_cur1}
\end{figure*}

\noindent \textbf{(F) Interest Generation:} 
We conducted a human annotation study focused on interest generation. For each dataset, 50 documents with manually annotated summaries were selected, and two independent annotators were asked to indicate whether the summary or the spotlight was more likely to generate interest for deeper exploration among readers. Annotators also had the option to state that both formats generate similar levels of interest. The overall inter-annotator agreement was 0.87. As shown in Column F of Table \ref{tab:qualitative-spot-sum}, annotators overwhelmingly rated spotlights as generating more interest than summaries. We find that the score differences (spotlight vs summary) are statistically significant ($p < 0.05$) as per McNemar's Test. 

The difference in reader engagement between a spotlight and a traditional summary is illustrated in Figure \ref{fig:spot_cur1}, which presents both generated from the same document describing a polygon approximation algorithm. The summary provides a concise technical account, focusing on adaptive vertex relocation and its effectiveness in minimizing approximation error, supported by experimental validation.
In contrast, the spotlight highlights the broader implications, potential impact, and limitations of the work, rather than delving into technical details. This framing encourages deeper reader engagement by implicitly prompting questions and curiosity. Additional details are available in the Appendix \ref{appendix:inquisitiveness_aspect_spot}.





\subsection{Spotlight-Summary: Similar Features}  \label{spot_summ_sim}

Apart from the aforementioned distinctions, a spotlight shares similarities with a summary in terms of faithfulness to the source document, length, and its qualification as a mini-story. This is detailed next. 

\noindent \textbf{(G) Factuality:} 
To quantitatively evaluate the faithfulness of condensations to their source documents, we employ FactScore \cite{factscore-2023}, a metric designed to evaluate factual consistency by extracting and aligning key information units. Using an information extraction approach, we extract structured factual triples from both the source document and its corresponding condensation. We then leverage a Natural Language Inference (NLI) model to determine whether each extracted fact from the condensation is entailed by the document. The FactScore is computed as the proportion of facts in the condensation that are supported by the source document. Our results in Table \ref{tab:qualitative-spot-sum} indicate that both summaries and spotlights achieve high FactScores, demonstrating strong factual alignment with the source content.

\noindent \textbf{(H) Length:} We checked the length (number of words) of both summary and spotlights and found that roughly they are of similar length.

\noindent \textbf{(I) Mini-story:} A mini-story is a short, focused narrative that conveys the core plot or theme of an article in a clear and engaging way. It summarizes key ideas or events concisely while preserving coherence and capturing the reader’s interest. Spotlight and summary contents for 25 articles of each dataset have been provided to two annotators to mark one of the following three - (i) both spotlight and summary contents have mini-story; (ii) only spotlight has mini-story; (iii) only summary has mini-story. Inter-Annotator agreement is 81.2\%. 
Any conflict has been resolved by another annotator. We find that for all the datasets at least 80\% of respondents have certified both the condensations as mini-story (spotlight has a slight higher rating). The detail is presented in Table \ref{tab:appendix_mini_story_prolific} (Appendix). 

\subsection{Spotlight-Highlight: Distinct Features}

To explore the distinctness between spotlight and highlights, we have performed a similar mini-story study to Section \ref{spot_summ_sim}. Details in Appendix Section \ref{prolific_mini_story}. In our analysis of distinct features (as described in Section 3.2), we observed that spotlights consistently outperform highlights in conveying a mini-story narrative. Annotator evaluations indicate that spotlights were rated as having a more engaging narrative as shown in Table \ref{tab:mini_story_spot_vs_highlight}. While both spotlights and highlights aim to distill key information from source documents, spotlights uniquely present a self-contained narrative that not only summarizes but also sparks curiosity and encourages further exploration. This narrative element is less pronounced in traditional highlights, which tend to list isolated key points rather than weaving them into a cohesive mini-story.

\begin{table}[]
    \centering
    \begin{adjustbox}{width=0.80\linewidth}
    \begin{tabular}{|c|c|c|}\hline
        \textbf{Dataset} & \textbf{Spotlight(\%)} & \textbf{Highlights(\%)} \\\hline
        News & 80 & 20\\\hline
        CSPubSum & 72 & 28\\\hline
        Wikipedia & 72 & 28\\\hline
        Research Presentation & 84 & 16\\\hline
    \end{tabular}
    \end{adjustbox}
    \caption{Mini-story validation for spotlight vs. highlight analysis across all datasets}
    \vspace{-3mm}
    \label{tab:mini_story_spot_vs_highlight}
\end{table}


\color{black}

\section{Spotlight Generation Process} 
\label{sec:spotGenPr}



For the task of spotlight generation, we propose a 2-step solution  - at the first step, we fine-tune an open-sourced LLM; in the next step, we apply direct preference optimization (DPO)~\cite{rafailov2024direct} framework on top of that fine-tuned LLM to generate  spotlights.

In the DPO approach, the model differentiates spotlights from summaries, and learns to prioritize spotlight generation. As discussed in previous sections, while summaries and spotlights share similarities in terms of length, faithfulness, and their mini-story nature, they differ in key aspects such as information distribution, compactness, and readability. By emphasizing these distinctions, the model enhances the uniqueness of spotlights, ultimately leading to higher-quality outputs.

\noindent \textbf{Dataset Preparation:} For each document, we consider the  ground truth spotlight as an upvoted example (indicating a preferred completion), while the corresponding
summary as a downvoted example (denoting undesirable).
Thus each document from the original dataset contributes one upvoted and one downvoted example.

\noindent\textbf{Experiment Steps:} We randomly divide each dataset into two equal parts for a two-step process. In the first step, one half is used for Supervised Fine-Tuning (SFT) on a base model (e.g., Llama2-7b-hf, Mistral-7B-v0.1, Llama3-8b) using ground truth spotlights. In the second step, the remaining half is utilized for Direct Preference Optimization (DPO) on the fine-tuned model obtained from the first step. In this stage, both spotlight and summary information are leveraged to highlight the distinctions necessary for the DPO process. 

\noindent \textbf{ Direct Preference Optimization:}
For each example in the second step, we consider the ground truth spotlight as the desired (preferred/chosen/winning) completion and the generic summary as the undesirable (rejected/losing) completion. 

 \textbf{Loss Function:} The DPO loss function is:
\vspace{-1mm}
\begin{align*}
     \mathcal{L}_{DPO} (\pi_{\theta};\pi_{ref})  = & 
     - \mathbf{E}_{(x,y_w,y_l)\sim \mathcal{D}}  \\ \Big{[}log \, \sigma \, \Big{(}\beta log\frac{\pi_\theta(y_w \, | x)}{\pi_{ref}(y_w \, | x)}
    & - \beta log\frac{\pi_\theta(y_l \, | x)}{\pi_{ref}(y_l \, | x)}\Big{)}\Big{]}
\end{align*}
where \(\pi_\theta\) is the language model we are optimizing, \(\pi_{ref}\) is the reference model (the fine-tuned model obtained after first step) with respect to which we are increasing the likelihood of preferred responses and decreasing the likelihood of nonpreferred responses. 
Samples in the preference dataset \(\mathcal{D}\) look like \((x,y_w,y_l)\) where \(x\) is the prompt (instruction with document in our case), \(y_w\) is the desired response (spotlight in our case) and \(y_l\)  is the undesired response (summary in our case). \(\beta\) is a tunable hyperparameter.

\noindent \textbf{Open Source LLMs:} We use various small-size large language models (LLMs) - Mistral-7b~\cite{jiang2023mistral} and Llama\footnote{\url{https://ai.meta.com/llama/}} based - Llama2-7b and Llama3-8b models.
We experiment with Mistral-7b (abbreviated as Mst-7b-FT), Llama2-7b and Llama3-8b fine tuning (abbreviated as L2-7b-FT and L3-8b-FT respectively), general instruction tuned vanilla models (Mst-7b-VA, L2-7b-VA and L3-8b-VA respectively) and corresponding DPO models. We report the best performing Llama3-8b LLM-based outcomes in Table~\ref{tab:rouge_gpt} and other details in the Appendix Tables \ref{tab:appendix-news_rouge_gpt}, \ref{tab:appendix-CSPubSum_rouge_gpt}, \ref{tab:appendix-oasum_rouge_gpt} and \ref{tab:appendix-neurips_rouge_gpt}.

\section{Experimental Setup}

For spotlight generation task, we experiment with several methods and analyze the generated spotlights based on various evaluation frameworks.  

\subsection{Baseline Approaches}
\label{approach-algo}

We consider two different approaches for baseline. 

\noindent \textbf{(A) Traditional Models:} We employ Longformer~\cite{beltagy2020longformer} 
(LF)
, 
T5-3b finetuned~\cite{raffel2020exploring} (T5-FT), BART-large~\cite{lewis2019bart}, instruction-tuned Pegasus~\cite{zhang2020pegasus}, Falcon 7b-instruction tuned model~\cite{penedo2023refinedweb} and state-of-the-art extreme summarization approach 
(TLDR)~\cite{cachola2020tldr} as baselines. 



\noindent\textbf{(B) Large-Size LLMs:} We apply example-based prompting (1-shot) on large-sized LLMs like ChatGPT\footnote{\url{https://chat.openai.com/}},
GPT-4\footnote{\url{https://openai.com/gpt-4}} and Gemini-Pro\footnote{\url{https://gemini.google.com/}} to check their capability to generate good quality spotlights. Due to space shortage, ChatGPT, and Gemini-Pro evaluations are in Appendix Tables \ref{tab:appendix-news_rouge_gpt}, \ref{tab:appendix-CSPubSum_rouge_gpt}, \ref{tab:appendix-oasum_rouge_gpt} and  \ref{tab:appendix-neurips_rouge_gpt}. 

\subsection{Evaluation} 
 We assess the quality of the generated spotlights using
the metrics defined in Table \ref{tab:qualitative-spot-sum}. Besides this we measure its \underline{similarity with the reference spotlights} by the following three set of evaluation metrics.

\noindent \textbf{(A) Traditional:} We check the competence of different models with traditional evaluations metrics like (i) Rouge1 (R1), Rouge2 (R2), 
~\cite{lin2004rouge} and 
(ii) BERT-F1 (BeF1)~\cite{zhang2019bertscore}.


\noindent \textbf{(B) UniEval~\cite{zhong2022towards}:} 
UniEval correlates 
much better with human judgments in 
NLG tasks. We use the `Relevance' of UniEval scores to identify the quality of spotlight generation.

\noindent \textbf{(C) GPT4 Critique~\cite{mullick2024persona}:}   
GPT-4 is asked to rate the spotlight – whether it is good or bad with respect to the ground truth spotlight and whether it generates interest to read the source document. 

\subsection{Experimental Settings} We use 80GB A100 GPU, 210MHz clock cycle with Huggingface TRL and PEFT libraries built on top of Huggingface Transformers to conduct the experiments of fine-tuning and alignment of pretrained models and for their evaluation purpose. Additionally, we use the NLTK, spaCy, and NumPy packages for further analysis and ablations. Further details are in Appendix \ref{appendix: experiments}.

\color{black}

\section{Experimental Results and Discussion}

\begin{table}[htb]
    \centering
    \begin{adjustbox}{width=0.90\linewidth}
    \begin{tabular}{|c|c|c|c|c|c|}
        \hline
        \multirow{2}{*}{\textbf{Model}} &  \multicolumn{3}{|c|}{\textbf{Traditional Metrics}} &  \multicolumn{1}{|c|}{\textbf{UniEval}} &  \multicolumn{1}{|c|}{\textbf{GPT-4}}\\
        \cline{2-5}
          & \textbf{R1} & \textbf{R2} & \textbf{BeF1} &  \textbf{Relevance} & \textbf{Critique}\\
        \hline
L2-7b-FT & 28.1 & 8.6 & 81.3 & 66.6 & 77.8\\\hline
Mst-7b-FT & 25.7 & 6.9 & 77.4 & 65.1 & 77.0\\\hline
L3-8b-FT & \textbf{31.9} & \textbf{9.7} & \textbf{81.8} & \textbf{67.1} & \textbf{79.5}\\\hline \hline
LF & 16.0 & 4.8 & 67.4 & 55.2 & 68.7
\\\hline
T5-FT & 20.0 & 5.0 & 71.0 & 
63.2 & 75.9\\\hline
BART & 19.1 & 3.3 & 71.9 & 
64.7 & 70.2\\\hline
Pegasus	& 17.8 & 4.1 & 70.8 & 
62.5 & 60.9\\\hline
Falcon& 17.0 & 3.0 & 74.0 & 
55.2 & 38.4\\\hline
TLDR & 19.5 & 5.6 & 77.6 & 62.9 & 72.3\\\hline
    \end{tabular}
    \end{adjustbox}
    \vspace{-2mm}
    \caption{Traditional, UniEval and GPT-4 Critique evaluations on CSPubSum Dataset - comparison of baselines with SFT, Additional Results in Appendix Table \ref{tab:news_rouge_gpt_baseline}, \ref{tab:cspubsum_rouge_gpt_baseline}, \ref{tab:oasum-domain_rouge_gpt_baseline} and \ref{tab:neurips_rouge_gpt_baseline}}
    \vspace{-3mm}
    \label{tab:cspubsum_rouge_gpt}
\end{table}

In this section, we analyze the results 
based on values of different evaluation strategies - traditional metrics, UniEval, GPT-4 critiquing and spotlight characteristics features to understand how different models perform and gain insights into the effectiveness of LLMs for spotlight generation tasks.

\noindent \textbf{(A) Performance of Traditional Baseline Approaches:} 
We explore different traditional approaches (LongFormer, T5-Fine-tune, BART-large, Pegasus, Falcon, TLDR) for spotlight generation and results for CSPubSum are shown in Table \ref{tab:cspubsum_rouge_gpt}. 
Table \ref{tab:cspubsum_rouge_gpt} shows that traditional summarization models such as LongFormer, T5, BART, Pegasus, and Falcon struggle to generate effective spotlights, with lower ROUGE, BERT-F1, and UniEval scores. In contrast, even simple supervised fine-tuning of LLMs (Llama2, Mistral, Llama3) yields large improvements, confirming that LLMs can adapt well to the spotlight objective. This gap underscores that generating spotlights demands richer semantic modeling than what traditional summarization frameworks provide. 
Detailed results for the datasets are in Appendix Table - \ref{tab:news_rouge_gpt_baseline} (for News), \ref{tab:cspubsum_rouge_gpt_baseline} (for CSPubSum), \ref{tab:oasum-domain_rouge_gpt_baseline} (for Wikipedia) and \ref{tab:neurips_rouge_gpt_baseline} (for Research Presentation) 
It is seen that even simple SFT-based approaches on LLMs perform much better than these traditional approaches across all metrics. This performance gap may be due to the advanced architectures and targeted optimizations of LLM models which are fine-tuned specifically for task performance and user preferences, whereas the older models do not benefit from the same level of refinement and cutting-edge techniques, leading to their comparatively lower performance. 

\begin{table}[htt]
    \centering
    \vspace{-2mm}
    \begin{adjustbox}{width=0.95\linewidth}
    \begin{tabular}{|c|c|c|c|c|c|}
        \hline
         \multirow{2}{*}{\textbf{Model}} &  \multicolumn{3}{|c|}{\textbf{Traditional Metrics}}  & \multicolumn{1}{|c|}{\textbf{UniEval}}&  \multicolumn{1}{|c|}{\textbf{GPT-4}}\\
        \cline{2-6}
        & \textbf{R1} & \textbf{R2} & \textbf{BeF1} &  \textbf{Relevance} & \textbf{Critique}\\
        \hline
\multicolumn{6}{|c|}{\textbf{News}}\\\hline
L3-8b-VA& 15.9 & 6.8 & 73.8 & 70.6 & 78.0\\\hline
GPT-4 (1-shot) & 31.3 & 12.1 & 79.5 & 83.1 & 90.1\\\hline
L3-8b-FT& 36.5 & 15.8 & 81.5 & 82.2 & 89.4\\\hline
DPO (L3-8b)& \textbf{36.8} & \textbf{16.0} & \textbf{83.6} & \textbf{84.7} & \textbf{91.3}\\\hline
\multicolumn{6}{|c|}{\textbf{CSPubSum}}\\\hline
L3-8b-VA& 15.5 & 3.8 & 47.2 & 60.1 & 73.6\\\hline
GPT-4 (1-shot) & 28.7 & 7.4 & 80.8 & 66.8 & 82.1\\\hline
L3-8b-FT & 31.9 & 9.7 & 81.8 &  67.1 & 79.5\\\hline
DPO (L3-8b)& \textbf{32.3} & \textbf{9.9} & \textbf{82.3} & \textbf{67.8} & \textbf{89.0}\\\hline
 \multicolumn{6}{|c|}{\textbf{Wikipedia}}\\\hline
L3-8b-VA & 20.3 & 6.5 & 71.7 & 65.7 & 52.1\\\hline
GPT-4 (1-shot) & 34.6 & 13.8 & 80.4 & \textbf{70.2} & \textbf{73.5}\\\hline
L3-8b-FT& 42.4 & 26.0 & 81.3 &  69.2 & 64.8\\\hline
DPO (L3-8b)& \textbf{42.5} & \textbf{26.3} & \textbf{83.4} & 67.3 & 67.3\\\hline
\multicolumn{6}{|c|}{\textbf{Research Presentation}}\\\hline
L3-8b-VA& 15.6 & 3.2 & 48.5 &  70.2 & 73.6\\\hline
GPT-4 (1-shot)& 32.3 & 7.1 & 77.2 & 75.2  & 81.6\\\hline
L3-8b-FT & 28.3 & 6.9 & 76.5 & 
 76.7 & 81.2\\\hline
DPO (L3-8b)& \textbf{38.2} & \textbf{8.9} & \textbf{79.6} & \textbf{76.8} & \textbf{88.8} \\\hline
    \end{tabular}
    \end{adjustbox}
    \vspace{-2mm}
    \caption{Traditional, UniEval and GPT-4 Critique evaluations for best performing Llama3-8b and GPT-4 approach in \% (best outcome in \textbf{Bold}) for each dataset, Additional evaluation details including other LLMs are in Appendix Table \ref{tab:appendix-news_rouge_gpt}, \ref{tab:appendix-CSPubSum_rouge_gpt}, \ref{tab:appendix-oasum_rouge_gpt} and \ref{tab:appendix-neurips_rouge_gpt}}
    \vspace{-2mm}
    \label{tab:rouge_gpt}
\end{table}




\begin{table}[!b]
\vspace{-3mm}
    \centering
    \begin{adjustbox}{width=\linewidth}
    \begin{tabular}{|c|c|c|c|c|c|c|}
        \hline
        \multirow{3}{*}{\textbf{Model}} &  \multicolumn{2}{|c|}{\textbf{Distribution}} & \multirow{1}{*}{\textbf{Compa-}} &  \multicolumn{3}{|c|}{\textbf{Readability(\%)}} \\
        \cline{2-3}\cline{5-7}
        & \multirow{2}{*}{Entropy$\downarrow$} & \multirow{2}{*}{Skew50$\uparrow$} & \textbf{ctness$\downarrow$} & Flesch & Flesch & \multirow{2}{*}{Gunning$\downarrow$} \\
        &  &  & & Kincaid$\downarrow$ &  Reading$\uparrow$ &  \\
        \hline
\multicolumn{7}{|c|}{\textbf{News}}\\\hline
L3-8b-FT& 0.65 & 0.51 & 0.62 & 9.09 & 62.39 & 10.67 \\\hline
DPO (L3-8b) & 0.60 & 0.55 & 0.57 & 9.01 & 64.83 & 10.63 \\\hline
\multicolumn{7}{|c|}{\textbf{CSPubSum}}\\\hline
L3-8b-FT& 0.60 & 0.53 & 0.65 & 9.38 & 59.36 & 10.72 \\\hline
DPO (L3-8b)& 0.57 & 0.50 & 0.61 & 9.30 & 62.96 & 10.14 \\\hline
\multicolumn{7}{|c|}{\textbf{Wikipedia}}\\\hline
L3-8b-FT&  0.63 & 0.55 & 0.71 & 7.41 & 66.31 & 8.67 \\\hline
DPO (L3-8b)  & 0.57 & 0.59 & 0.65 & 6.80 & 67.65 & 8.01\\\hline
\multicolumn{7}{|c|}{\textbf{Research Presentation}}\\\hline
L3-8b-FT & 0.49 & 0.40 & 0.37 & 12.0 & 44.74 & 12.12 \\\hline
DPO (L3-8b) & 0.45 & 0.47 & 0.30 & 9.62 & 55.85 & 9.57 \\\hline
    \end{tabular}
    \end{adjustbox}
    \vspace{-2mm}
    \caption{Spotlight Characteristics Evaluation on Model Generated spotlight -  Entropy, Compactness, Flesch Kincaid and Gunning score lower values signify better result, but for Skew50 and Flesch Reading score higher values signify a better result as discussed in Section \ref{spotlight-define} }
     \vspace{-2mm}
    \label{tab:spot_char_model}
\end{table}

\noindent \textbf{(B) Performance of DPO vis-a-vis other LLM-based approaches:} The results presented in Table \ref{tab:rouge_gpt}, demonstrate the effectiveness of the DPO approach for spotlight generation across various evaluation criteria for Llama3-8b large language model (LLM) for different datasets where L3-8b-VA, L3-8b-FT and DPO (L3-8b) indicate Llama3-8b vanilla, supervised fine-tuning and DPO approach respectively.
It is seen that 
\begin{itemize}
    \item DPO consistently outperforms vanilla models, supervised fine-tuning (SFT), and GPT-4 (1-shot) (in most of the cases) 
proving its superiority in spotlight generation.
    \item SFT approaches is a close second, however, DPO consistently improves over SFT methods  indicating its utility and  the potency of SFT when the reference negative examples (here summaries) are not available.
    \item Vanilla L3-8b-VA performs the worst, struggling in both traditional and advanced evaluations, highlighting the importance of fine-tuning or DPO across all datasets.
    \item for \textit{News, CSPubSum and Research Presentation} datasets, Llama3-8b fine-tuning and DPO, both perform better than GPT-4 (1-shot) for traditional metrics whereas for \textit{Research Presentation}, GPT-4 (1-shot) performs better than the fine-tuning model but worse than DPO.
    \item GPT-4 (1-shot) is strong in UniEval Relevance and Critique (it is best among all in \textit{Wikipedia}) but falls behind in ROUGE and BERTScore, suggesting it produces more readable but not necessarily more extractive spotlights. 
    \item For Wikipedia data, all models show higher Rouge scores with lower UniEval relevance/GPT-4 Critique suggests that while the spotlights have significant lexical overlap with the reference, they may lack deeper semantic coherence.
    \item We explore different DPO configurations for utilization of data and fine-tuning approach (details in Appendix \ref{app:dpo-variations}) and the best result is obtained in our 2-step approach when half of the data is used for SFT followed by DPO for the remaining half of the data.
    \item Notably, the Llama3-8b based DPO model achieves statistically significant improvements (p < 0.05) over other model variations. 
    \item To validate the GPT-4 critique scores against robust human judgment, we conducted an additional human annotation study similar to \cite{mullick2024persona} to evaluate the alignment between human judgment and automated assessments for 20 documents for each dataset, details in Appendix \ref{appendix:dpo_comparison_and_discussion}. We obtain an overall average Point-biserial correlation coefficient of 0.751 between the GPT-4 critique scores and human judgment. 
\end{itemize}
Across the datasets, DPO applied on Llama3-8b consistently outperforms both vanilla and fine-tuned versions. While GPT-4 (1-shot) is strong in certain cases, particularly in  UniEval, the DPO-trained Llama3-8b achieves the best balance across all metrics. This demonstrates that direct preference optimization is especially effective for spotlight generation.

\color{black}

\noindent \textbf{(C) DPO Conformance to Spotlight Features:}
We do an extensive evaluation of the model-generated spotlights of DPO pipeline with the fine-tuned version 
across various spotlight characteristics, as detailed in Table \ref{tab:spot_char_model} which shows that
(i) DPO (L3-8b) consistently improves readability across all datasets, producing spotlights that are easier to understand than fine-tune approach; (ii) Entropy reductions and increase in Skew50  across all datasets imply DPO maintains skewness in term distribution than fine-tune. (iii) Compactness decreases in DPO, indicating less information overload and more readability. 
In conclusion, DPO produces more structured, readable, and well-balanced spotlights compared to fine-tuned approach.
These observations confirm that DPO yields spotlights that are not only more structured and readable but also more faithful to the intended characteristics of spotlight generation, in contrast to the less targeted outputs of a purely fine-tuned approach.

\begin{table}[]
    \centering
    \begin{adjustbox}{width=0.99\linewidth}
    \begin{tabular}{|c|c|c|c|c|c|c|c|c|c|c|}
        \hline
        \multirow{2}{*}{\textbf{Model}} &  \multicolumn{3}{|c|}{\textbf{Traditional Metrics}} &  \multicolumn{1}{|c|}{\textbf{UniEval}} & \multicolumn{1}{|c|}{\textbf{GPT-4}}\\
        \cline{2-5}
        & \textbf{R1} & \textbf{R2} & \textbf{BeF1} & \textbf{Relevance} & \textbf{Critique}\\
        \hline
\multicolumn{6}{|c|}{\textbf{News}} \\\hline
L3-8b-VA & 17.2 & 8.3 & 75.8 & 72.5 & 65.0\\\hline
L3-8b-FT Spot$\dagger$ & 24.6 & 12.5 & 75.0 & 76.3 & 61.3\\\hline
DPO (L3-8b) Spot$\dagger$ & 23.6 & 10.6 & 70.7 &  74.9 & 60.8\\\hline
L3-8b-FT Sum$\mathsection$ & \textbf{57.3} & \textbf{36.0} & \textbf{85.2} &  \textbf{81.9} & \textbf{87.3}\\\hline
\multicolumn{6}{|c|}{\textbf{CSPubSum}} \\\hline
L3-8b-VA & 13.5 & 2.5 & 60.2 & 61.8 & 59.6\\\hline
L3-8b-FT Spot$\dagger$ & 12.5 & 1.4 & 61.0 & 63.9 & 65.3\\\hline
DPO (L3-8b) Spot$\dagger$ & 10.1 & 1.8 & 59.8 & 62.5 & 60.8\\\hline
L3-8b-FT Sum$\mathsection$ & \textbf{25.0} & \textbf{7.5} & \textbf{70.1} & \textbf{66.5} & \textbf{73.0}\\\hline
\multicolumn{6}{|c|}{\textbf{Wikipedia}}\\\hline
L3-8b-VA & 19.2 & 7.8 & 65.7 & 66.1 & 55.1\\\hline
L3-8b-FT Spot$\dagger$ & 28.3 & 13.8 & 64.7 & 72.5 & 48.1\\\hline
DPO (L3-8b) Spot$\dagger$ & 25.9 & 12.5 & 60.7 & 70.3 & 45.3\\\hline
L3-8b-FT Sum$\mathsection$ & \textbf{32.3} & \textbf{18.2} & \textbf{77.3} & \textbf{78.6} & \textbf{70.2}\\\hline
\multicolumn{6}{|c|}{\textbf{Research Presentation}}\\\hline
L3-8b-VA & 14.8 & 3.8 & 58.5 & 69.5 & 63.6\\\hline
L3-8b-FT Spot$\dagger$ & 32.5 & 12.1 & 70.1 & 75.3 & 68.9\\\hline
DPO (L3-8b) Spot$\dagger$ & 29.3 & 7.1 & 68.8 & 71.8 & 61.9\\\hline
L3-8b-FT Sum$\mathsection$ & \textbf{38.3} & \textbf{16.2} & \textbf{74.7} & \textbf{77.4} & \textbf{75.3}\\\hline
    \end{tabular}
    \end{adjustbox}
    \vspace{-2mm}
    \caption{Traditional and GPT-4 Critique Evaluations for each dataset, $\dagger$: model trained on spotlights but evaluated on gold standard summary (GPT-4 generated), $\mathsection$: model trained on summary and evaluated on gold standard summary}
     \vspace{-5mm}
    \label{tab:rouge_gpt_summ_gold}
\end{table}

\noindent \textbf{(D) Divergence of DPO-generated Spotlights from the gold standard Summary:} 
As outlined in Section~\ref{sec:spotGenPr}, our spotlight generation process consists of two steps. First, we fine-tune the base model (FT), and then we apply the DPO method to the fine-tuned model to align the output closer to a spotlight and further from a summary. While DPO produces superior spotlights (Table~\ref{tab:rouge_gpt}), a question remains: does the spotlight diverge from the summary characteristics, as per the DPO objective?
To answer this, we compare model-generated spotlights with gold-standard summaries (GPT-4 generated) for three approaches: vanilla (L3-8b-VA), fine-tuning (L3-8b-FT Spot), and DPO (DPO (L3-8b) Spot), as presented in Table~\ref{tab:rouge_gpt_summ_gold} for the entire datasets. Our findings indicate that the spotlights generated using SFT are closer to the reference summaries than those produced by DPO, confirming the deviation. 
Table ~\ref{tab:rouge_gpt_summ_gold} examines whether spotlight-trained models drift toward summary characteristics. The results indicate that spotlights produced by SFT are closer to gold summaries, whereas DPO outputs diverge more strongly, aligning with the intended spotlight objective. At the same time, when models are trained explicitly on summaries, they achieve higher traditional metrics but lose the engagement qualities of spotlights. This confirms that while spotlights and summaries overlap, optimizing for one does not automatically guarantee the qualities of the other. 
Nonetheless, both the fine-tuned and DPO outputs outperform the vanilla model, which prompts us to investigate whether the spotlight generation process inadvertently produces quality summaries.

To explore this further, we fine-tuned Llama3-8b with reference summaries (L3-8b-FT Sum) to generate concise responses. This approach consistently achieves significantly higher scores across traditional metrics (R1, R2, BeF1) as well as improved GPT-4 Critique and UniEval evaluations (Table~\ref{tab:rouge_gpt_summ_gold}).
These results suggest that, although there is some overlap between spotlights and summaries, the optimization for spotlights introduces only a partial summary flavor, while remaining distinct from a true summary. 

\color{black}
\section{Conclusion}

In this paper, we introduced the concept of “Spotlight,” a novel, concise representation of source documents that offers a different perspective to conventional summaries and highlights. Our work identifies a set of unique characteristics that differentiate spotlights, emphasizing their ability to present key aspects of a document in a faithful yet compelling way, effectively capturing the reader’s attention as well as drawing their attention to the source document.  
To validate this concept, we prepared four carefully annotated datasets, which served as the basis for a comprehensive analysis. Through rigorous quantitative evaluations and human studies, we established nine key features that differentiate spotlights primarily in terms of coverage and presentation style.
Building on these findings, we proposed a Direct Preference Optimization (DPO)-based fine-tuning algorithm to generate spotlights. Extensive evaluation using both traditional metrics and LLM-based assessments demonstrate the effectiveness of our approach. We believe that, similar to summarization, there is a significant potential for developing various types of spotlights — such as aspect-based, query-focused, and personalized versions — which will be the focus of our immediate future work.

\section*{Limitations}
Our datasets are neither multilingual nor multimodal. So, we need to explore how state-of-the-art approaches can be utilized in multilingual and multimodal scenarios for spotlight generation - which we aim to do as a part of future work. We have also explored some edge cases and limitations of the model generated spotlights as in Appendix \ref{appendix : spot_limitation}.

\section*{Ethics Statement}
Our work does not reveal any personal sensitive information and we use publicly available benchmarked datasets and models in different contexts.


\bibliography{custom}

\appendix

\newpage
\newpage

\section*{Appendix}

\section{Dataset} \label{data-oasum}

\begin{table}[!ht]
    \centering
    \begin{adjustbox}{width=0.98\columnwidth}
    \begin{tabular}{|c|c|}
        \hline
        Domain & Categories \\
        \hline
        HealthCare & Death, Diagnosis, Differential diagnosis, Diagnosis-Classification \\
        \hline
        Education & History, Geography, Taxonomy, Education\\
        \hline
        Life and Career & Career, Political Career, Personal Life, Life and career\\
        \hline
         Music & Production, Composition, Soundtrack, Track Listing\\
        \hline
    \end{tabular}
    \end{adjustbox}
    \caption{Domain-wise breakdown of Categories in Wikipedia dataset}
    \label{tab:domain_categories_set}
\end{table}

\begin{figure*}[!ht]
    \vspace{-1mm}
    \centering
    \includegraphics[width=0.97\textwidth]{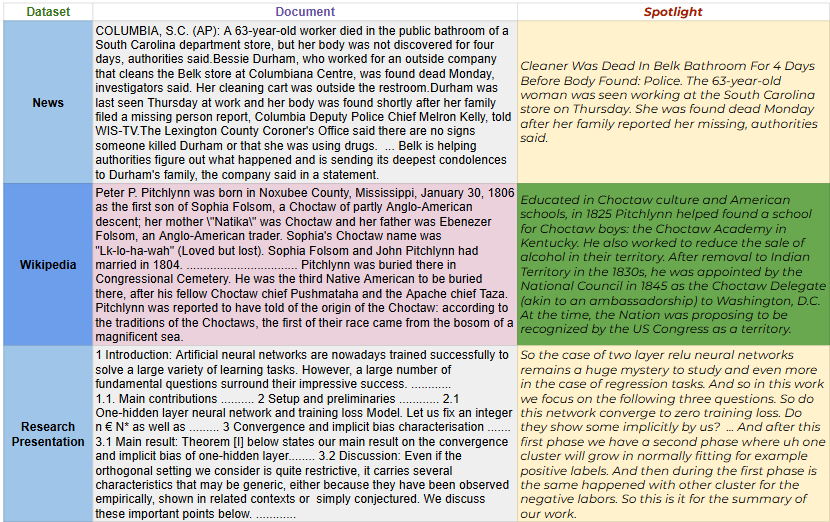}
    \caption{Examples of \underline{(Document, Spotlight)} pairs from different datasets, An example from CSPubSum is given in Figure \ref{fig:spot_cur1}, The documents and the example spotlight of Research Presentation have been truncated in this visualization owing to their length. Across the three examples, the spotlights consistently transform lengthy, detail-heavy texts into short, curiosity-driven narratives. In the News case, procedural details and background information are removed, leaving a striking account that emphasizes the unusual and attention-grabbing aspect of the story—the body discovered after four days. The Wikipedia example shifts from an encyclopedic biographical entry to a narrative that foregrounds milestones and memorable achievements, making the article more engaging and easier to recall. Finally, the Research Presentation example reframes dense, technical exposition into an accessible narrative by highlighting the central research question and its intrigue rather than methodological detail. Together, these examples illustrate how spotlights distill complex information into focused, compelling narratives that spark interest rather than merely summarize content.}
    \label{fig:spot_egs}
\end{figure*}

\begin{figure}[!ht]
    \centering
    \includegraphics[scale=0.7]{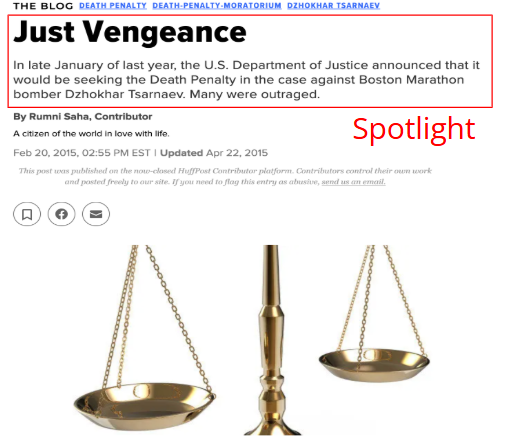}
    \caption{News Headline and the Short Description taken as spotlight for the News dataset. This figure illustrates the construction process of the News spotlight dataset. Source articles are paired with human-curated title and short description.}
    \label{fig:news_cat_spot}
\end{figure}

\noindent Figure \ref{fig:news_cat_spot}  demonstrate how we repurpose attributes of News dataset as spotlights.

\begin{figure}[!ht]
    \centering
    \includegraphics[scale=0.85]{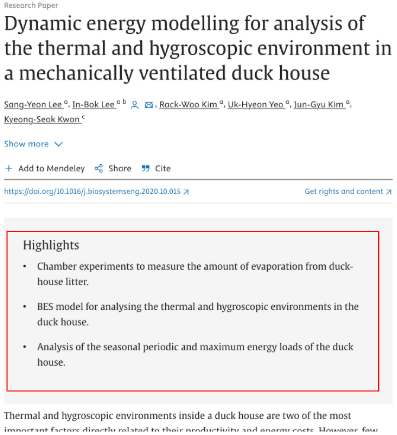}
    \caption{Author written highlights taken as spotlight for the CSPubSum dataset. From the Research Papers the author-written bullet points serve as a compelling entryway into the research paper, offering a concise preface that draws attention to the central questions and findings while setting an engaging narrative tone.}
    \label{fig:cspubsuym_spot}
\end{figure}

\begin{figure*}[!ht]
    \centering
    \includegraphics[scale=0.7]{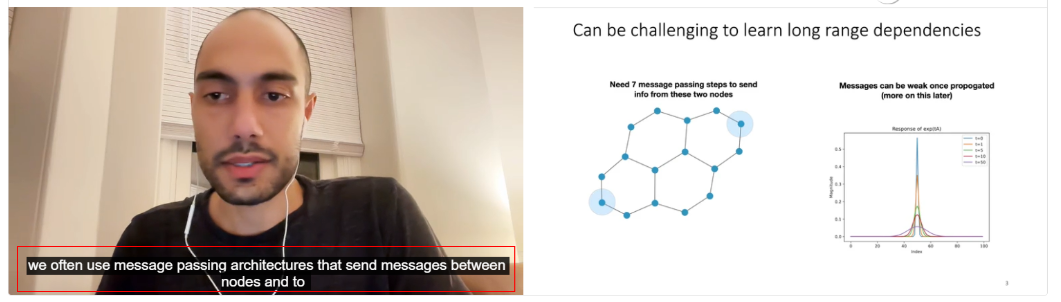}
    \caption{Conference Video Transcription taken as spotlight for the Research Presentation dataset. From slide decks and talk transcripts, concise spotlights are crafted to reflect the most engaging aspects of the research story. Unlike abstracts or slide summaries, these spotlights focus on intrigue, contributions, and memorable phrasing, ensuring that the dataset captures narrative appeal specific to academic communication.}
    \label{fig:research_ppt_spot}
\end{figure*}

\section{Spotlight: Characterization Details}

In this section, we further discuss different spotlight characteristics in detail. While comparing with summary, analysis has been done using two different types of summaries - Human Annotated (60 Articles of each dataset) and GPT-4 LLM generated (for all articles of each dataset). 

\begin{table}[th]
    \centering
    \begin{adjustbox}{width=0.9\linewidth}
    \begin{tabular}{|c|c|}
    \hline
        \textbf{Metric} & \textbf{Better Score} \\
        \hline
        Flesch-Kincaid~\cite{flesch1948new} & Lower ($\downarrow$)\\\hline
        Flesch Reading~\cite{flesch1960write} & Higher ($\uparrow$)\\\hline
        Gunning Index~\cite{gunning1952technique} & Lower ($\downarrow$)\\
        \hline
        Dale-Chall Score~\cite{dale1948formula} & Lower ($\downarrow$)\\\hline
        McLaughlin's SMOG Score~\cite{caylor1973methodologies} & Lower ($\downarrow$)\\\hline
        FORCAST ~\cite{caylor1973methodologies} & Higher ($\uparrow$)\\\hline
    \end{tabular}
    \end{adjustbox}
    \caption{Comparison of Readability Metrics}
    \label{tab:appendix_readability}
\end{table}

\subsection{Readability}
\label{app:read}

As shown in Table \ref{tab:appendix_readability}, we utilize six different readable metrics as following:

(i) \textit{Flesch-Kincaid Grade Level~\cite{flesch1948new}:} The Flesch-Kincaid Grade Level is a readability metric indicating the U.S. school grade level needed to understand a text, calculated from sentence length and syllable count, with lower scores signifying easier readability.\\
(ii) \textit{Flesch Reading Ease Score~\cite{flesch1960write}:} The Flesch Reading Ease Score is a readability metric, ranging from 0 to 100, that assesses text complexity based on sentence length and syllable count, with higher scores indicating easier readability. \\
(iii) \textit{Gunning Fog Index~\cite{gunning1952technique}:} The Gunning Fog Index is a readability metric estimating the years of formal education needed to understand a text on the first reading, with lower scores indicating easier readability and less educational requirement.\\ 
(iv) \textit{Dale-Chall Score~\cite{dale1948formula}:} The Dale-Chall Score is a readability metric assessing text complexity based on sentence length and the percentage of difficult words, with lower scores indicating easier-to-read texts that use common words and shorter sentences, making them accessible to a broader audience.\\ 
(v) \textit{McLaughlin's SMOG Score~\cite{caylor1973methodologies}:} The SMOG (Simple Measure of Gobbledygook) formula estimates the readability of a text by assessing the number of polysyllabic words in a set number of sentences, with lower scores indicating easier readability and requiring fewer years of education to understand.\\ 
(vi) \textit{FORCAST Formula Score~\cite{caylor1973methodologies}:} The FORCAST formula assesses text complexity, especially in technical and scientific documents, by counting single-syllable words in a 150-word sample, with higher scores indicating easier readability contrary to many readability formulas.\\ 

Table \ref{tab:qualitative-spot-sum-human} shows complete comparative evaluations of six different readability metrics for spotlight (Spot) vs. Human Generated summary (Summ). Similarly, comparative evaluations of six different readability metrics for spotlight (Spot) vs. GPT-4 summary (Summ) are presented in Table \ref{tab:qualitative-spot-sum-gpt4}. Both the Tables infer that all readability metrics indicate that spotlight is more readable than summary, similar to Table \ref{tab:qualitative-spot-sum} assessment.


\begin{table*}[!ht]
\vspace{1mm}
\centering
\begin{adjustbox}{width=\linewidth}
\begin{tabular}{|c|c|c|c|c|cccccc|c|}
\hline
\multicolumn{2}{|c|}{Dataset} & \multicolumn{2}{c|}{Distribution}                                                                               & Compact- & \multicolumn{6}{c|}{Readability (\%)}   &  Interest-  \\\cline{3-4} \cline{6-11}
                  \multicolumn{2}{|c|}{} & \multirow{2}{*}{Entropy$\downarrow$} & \multirow{2}{*}{Skew50$\uparrow$} &  ness$\downarrow$ & \multicolumn{1}{c|}{Flesch} & \multicolumn{1}{c|}{Flesch} & \multirow{1}{*}{Gunning $\downarrow$} & \multicolumn{1}{|c|}{Dale} & \multicolumn{1}{c|}{McLaug} & FORCAST$\uparrow$ &           Generation (\%)$\uparrow$\\
                  \multicolumn{2}{|c|}{} &  &  &   & \multicolumn{1}{c|}{Kincaid$\downarrow$} & \multicolumn{1}{c|}{Reading$\uparrow$} &  & \multicolumn{1}{|c|}{Chall$\downarrow$} & \multicolumn{1}{c|}{hlin's$\downarrow$} &  &  \\\hline \hline
              \multirow{2}{*}{News} & Spot  & \multicolumn{1}{c|}{0.69}                  & \multicolumn{1}{c|}{0.61}                            &      0.52             & \multicolumn{1}{c|}{10.57}                  & \multicolumn{1}{c|}{15.46}                  & \multicolumn{1}{c|}{16.81}                  & \multicolumn{1}{c|}{13.21}                  & \multicolumn{1}{c|}{20.69}                  &    1.96               &     80.0              \\ \cline{2-12}
               &  Summ    & \multicolumn{1}{c|}{0.96}                  & \multicolumn{1}{c|}{0.02}                              &          1.53         & \multicolumn{1}{c|}{16.89}                  & \multicolumn{1}{c|}{9.45}                  & \multicolumn{1}{c|}{20.10}                  & \multicolumn{1}{c|}{17.24}                  & \multicolumn{1}{c|}{26.10}                  &       -0.06            &     24.0              \\ \hline \hline
                \multirow{2}{*}{CSPubSum} & Spot  & \multicolumn{1}{c|}{0.62}                  & \multicolumn{1}{c|}{0.47}                                   &        0.62           & \multicolumn{1}{c|}{16.28}                  & \multicolumn{1}{c|}{9.67}                  & \multicolumn{1}{c|}{19.73}                  & \multicolumn{1}{c|}{13.21}                  & \multicolumn{1}{c|}{26.72}                  &      2.01             &        86.0           \\ \cline{2-12}
                & Summ   & \multicolumn{1}{c|}{0.96}                  & \multicolumn{1}{c|}{0.01}                                  &         1.28          & \multicolumn{1}{c|}{20.72}                  & \multicolumn{1}{c|}{6.42}                  & \multicolumn{1}{c|}{23.55}                  & \multicolumn{1}{c|}{17.74}                  & \multicolumn{1}{c|}{30.39}                  &    -0.08               &      30.0             \\ \hline\hline
              \multirow{2}{*}{Wikipedia}  & Spot    & \multicolumn{1}{c|}{0.76}                  & \multicolumn{1}{c|}{0.29}                        &      0.82             & \multicolumn{1}{c|}{12.62}                  & \multicolumn{1}{c|}{16.26}                  & \multicolumn{1}{c|}{16.37}                  & \multicolumn{1}{c|}{13.62}                  & \multicolumn{1}{c|}{23.32}                  &    1.36               &     86.0              \\ \cline{2-12}
                & Summ   & \multicolumn{1}{c|}{0.95}                  & \multicolumn{1}{c|}{0.05}                                 &        1.48           & \multicolumn{1}{c|}{16.24}                  & \multicolumn{1}{c|}{9.35}                  & \multicolumn{1}{c|}{19.19}                  & \multicolumn{1}{c|}{17.59}                  & \multicolumn{1}{c|}{27.35}                  &    0.04               &       26.0           \\ \hline \hline

              Research & Spot   & \multicolumn{1}{c|}{0.65}                  & \multicolumn{1}{c|}{0.36}                        &      0.28             & \multicolumn{1}{c|}{11.84}                  & \multicolumn{1}{c|}{44.10}                  & \multicolumn{1}{c|}{11.85}                  & \multicolumn{1}{c|}{13.82}                  & \multicolumn{1}{c|}{25.12}                  &    2.32               &     82.0              \\ \cline{2-12}
                Presentation & Summ   & \multicolumn{1}{c|}{0.92}                  & \multicolumn{1}{c|}{0.04}                                 &        0.69           & \multicolumn{1}{c|}{16.03}                  & \multicolumn{1}{c|}{23.83}                  & \multicolumn{1}{c|}{17.21}                  & \multicolumn{1}{c|}{17.23}                  & \multicolumn{1}{c|}{36.25}                  &         0.02          &        23.0          \\ \hline
\end{tabular}
\end{adjustbox}
\caption{Feature level analysis of spotlight (Spot) vs. human annotated summary (Summ) for 60 articles of each datasets, For Entropy, Compactness, Flesch Kincaid, Gunning score, Dale-Chall and McLaughlin's score lower values signifies better result, whereas for Skew50, Flesch Reading score, FORCAST and Interest-Generation higher values signifies a better result as discussed in Section \ref{spotlight-define}, More Exploratory Results of Table \ref{tab:qualitative-spot-sum}.
}
\vspace{-3mm}
\label{tab:qualitative-spot-sum-human}
\end{table*}

\begin{table}[htb]
\begin{adjustbox}{width=\linewidth}
\begin{tabular}{|c|c|cccccc|}
\hline
\multicolumn{2}{|c|}{\multirow{2}{*}{Dataset}} & \multicolumn{3}{|c|}{\textit{threshold=0.3}} &  \multicolumn{3}{|c|}{\textit{threshold=0.8}}                 \\ \cline{3-8} 
                  \multicolumn{2}{|c|}{} & \multicolumn{1}{c|}{R1} & \multicolumn{1}{c|}{R2} & \multicolumn{1}{c|}{RL} &  \multicolumn{1}{c|}{R1} & \multicolumn{1}{c|}{R2} & \multicolumn{1}{c|}{RL} \\\hline
                \multirow{2}{*}{News} & Spot  & \multicolumn{1}{c|}{0.910} & \multicolumn{1}{c|}{0.628} & \multicolumn{1}{l|}{ 0.521} &  \multicolumn{1}{c|}{0.021} & \multicolumn{1}{c|}{0.015} & \multicolumn{1}{c|}{0.016} \\ \cline{2-8}
                & Summ  & \multicolumn{1}{c|}{2.585} & \multicolumn{1}{c|}{1.185} & \multicolumn{1}{l|}{2.314}  & \multicolumn{1}{c|}{0.011} & \multicolumn{1}{c|}{0.010} & \multicolumn{1}{c|}{0.012} \\ \hline
                 \multirow{2}{*}{CSPubSum} & Spot & \multicolumn{1}{c|}{1.562} & \multicolumn{1}{c|}{0.362} & \multicolumn{1}{l|}{0.782} &  \multicolumn{1}{c|}{0.109} & \multicolumn{1}{c|}{0.025} & \multicolumn{1}{c|}{0.110} \\ \cline{2-8}
                 & Summ & \multicolumn{1}{c|}{3.433} & \multicolumn{1}{c|}{0.966} & \multicolumn{1}{l|}{2.566} & \multicolumn{1}{c|}{0.100} &  \multicolumn{1}{c|}{0.012} & \multicolumn{1}{c|}{0.083} \\ \hline
                \multirow{2}{*}{Wikipedia} & Spot  & \multicolumn{1}{c|}{1.826} & \multicolumn{1}{c|}{0.542} & \multicolumn{1}{c|}{1.528} &  \multicolumn{1}{c|}{0.032} & \multicolumn{1}{c|}{0.019} & \multicolumn{1}{c|}{0.029} \\ \cline{2-8}
                 & Summ & \multicolumn{1}{c|}{4.000} & \multicolumn{1}{c|}{1.233} & \multicolumn{1}{c|}{3.250}  & \multicolumn{1}{c|}{0.025} & \multicolumn{1}{c|}{0.015} & \multicolumn{1}{c|}{0.020} \\ \hline
                 Research & Spot  & \multicolumn{1}{c|}{1.431} & \multicolumn{1}{c|}{0.371} & \multicolumn{1}{c|}{0.836} &  \multicolumn{1}{c|}{0.052} & \multicolumn{1}{c|}{0.023} & \multicolumn{1}{c|}{0.024} \\ \cline{2-8}
                 Presentation & Summ & \multicolumn{1}{c|}{3.569} & \multicolumn{1}{c|}{1.025} & \multicolumn{1}{c|}{2.865}  & \multicolumn{1}{c|}{0.026} & \multicolumn{1}{c|}{0.018} & \multicolumn{1}{c|}{0.019} \\ \hline
\end{tabular}
\end{adjustbox}
\vspace{-2mm}
\caption{Feature level analysis on Informativeness and Degree of Extraction: spotlight (Spot) vs. Human Annotated summary (Summ), Additional Results of Table \ref{tab:qualitative-spot-sum}} 
\vspace{-5mm}
\label{tab:degree-infor-extract-human}
\end{table}

\begin{table*}[htb]
\vspace{-2mm}
\centering
\begin{adjustbox}{width=\linewidth}
\begin{tabular}{|c|c|c|c|c|cccccc|c|}
\hline
\multicolumn{2}{|c|}{Dataset} & \multicolumn{2}{c|}{Distribution}                                                                               & Compact- & \multicolumn{6}{c|}{Readability (\%)}   &  Interest-  \\\cline{3-4} \cline{6-11}
                  \multicolumn{2}{|c|}{} & \multirow{2}{*}{Entropy$\downarrow$} & \multirow{2}{*}{Skew50$\uparrow$} &  ness$\downarrow$ & \multicolumn{1}{c|}{Flesch} & \multicolumn{1}{c|}{Flesch} & \multirow{1}{*}{Gunning $\downarrow$} & \multicolumn{1}{|c|}{Dale} & \multicolumn{1}{c|}{McLaug} & FORCAST$\uparrow$ &           Generation (\%)$\uparrow$\\
                  \multicolumn{2}{|c|}{} &  &  &   & \multicolumn{1}{c|}{Kincaid$\downarrow$} & \multicolumn{1}{c|}{Reading$\uparrow$} &  & \multicolumn{1}{|c|}{Chall$\downarrow$} & \multicolumn{1}{c|}{hlin's$\downarrow$} &  &  \\\hline \hline
              \multirow{2}{*}{News} & Spot    & \multicolumn{1}{c|}{0.58}                  & \multicolumn{1}{c|}{0.43}                            &      0.68             & \multicolumn{1}{c|}{6.57}                  & \multicolumn{1}{c|}{65.86}                  & \multicolumn{1}{c|}{8.46}                  & \multicolumn{1}{c|}{10.58}                  & \multicolumn{1}{c|}{14.35}                  &    2.82               &     72.0              \\ \cline{2-12}
                &  Summ    & \multicolumn{1}{c|}{0.91}                  & \multicolumn{1}{c|}{0.33}                              &          0.78         & \multicolumn{1}{c|}{11.24}                  & \multicolumn{1}{c|}{49.31}                  & \multicolumn{1}{c|}{13.40}                  & \multicolumn{1}{c|}{11.42}                  & \multicolumn{1}{c|}{16.98}                  &       -0.90            &     72.0              \\ \hline
                \multirow{2}{*}{CSPubSum}  &  Spot  & \multicolumn{1}{c|}{0.66}                  & \multicolumn{1}{c|}{0.47}                                   &        0.50           & \multicolumn{1}{c|}{8.69}                  & \multicolumn{1}{c|}{49.67}                  & \multicolumn{1}{c|}{11.75}                  & \multicolumn{1}{c|}{10.61}                  & \multicolumn{1}{c|}{17.55}                  &      4.01             &        81.0           \\ \cline{2-12}
                &  Summ   & \multicolumn{1}{c|}{0.89}                  & \multicolumn{1}{c|}{0.22}                                  &       0.77            & \multicolumn{1}{c|}{14.31}                  & \multicolumn{1}{c|}{23.74}                  & \multicolumn{1}{c|}{16.42}                  & \multicolumn{1}{c|}{12.68}                  & \multicolumn{1}{c|}{20.40}                  &    -0.27               &     37.0             \\ \hline
              \multirow{2}{*}{Wikipedia}  & Spot    & \multicolumn{1}{c|}{0.76}                  & \multicolumn{1}{c|}{0.29}                        &      0.69             & \multicolumn{1}{c|}{8.87}                  & \multicolumn{1}{c|}{59.22}                  & \multicolumn{1}{c|}{10.97}                  & \multicolumn{1}{c|}{11.19}                  & \multicolumn{1}{c|}{16.12}                  &    1.15               &     83.0              \\ \cline{2-12}
                &  Summ   & \multicolumn{1}{c|}{0.88}                  & \multicolumn{1}{c|}{0.18}                                 &      0.89             & \multicolumn{1}{c|}{9.80}                  & \multicolumn{1}{c|}{52.99}                  & \multicolumn{1}{c|}{11.95}                  & \multicolumn{1}{c|}{12.36}                  & \multicolumn{1}{c|}{17.35}                  &    1.04               &       29.0            \\ \hline
               Research  &  Spot    & \multicolumn{1}{c|}{0.62}                  & \multicolumn{1}{c|}{0.38}                        &      0.26             & \multicolumn{1}{c|}{11.64}                  & \multicolumn{1}{c|}{38.10}                  & \multicolumn{1}{c|}{13.55}                  & \multicolumn{1}{c|}{11.80}                  & \multicolumn{1}{c|}{25.10}                  &    1.51               &     75.0              \\ \cline{2-12}
               Presentation   & Summ   & \multicolumn{1}{c|}{0.84}                  & \multicolumn{1}{c|}{0.18}                                 &        0.72           & \multicolumn{1}{c|}{16.23}                  & \multicolumn{1}{c|}{24.23}                  & \multicolumn{1}{c|}{17.81}                  & \multicolumn{1}{c|}{15.43}                  & \multicolumn{1}{c|}{32.65}                  &         0.05          &        25.0          \\ \hline
\end{tabular}
\end{adjustbox}
\caption{Feature level Analysis of spotlight (Spot) vs. GPT-4 generated summary (Summ) for entire datasets, For Entropy, Compactness, Flesch Kincaid, Gunning score, Dale-Chall and McLaughlin's score lower values signifies better result, whereas for Skew50, Flesch Reading score, FORCAST and Interest-Generation higher values signifies a better result as discussed in Section \ref{spotlight-define}, Additional Results of Table \ref{tab:qualitative-spot-sum}}
\label{tab:qualitative-spot-sum-gpt4}
\end{table*}

\subsection{Informativeness and Degree of Extraction}

Exploring further, we add Rouge-2 (R2) and Rouge-L (RL) evaluations for the Informativeness and Degree of Extraction along with the Rouge-1 (R1) score as discussed already in the main paper in Section \ref{spot-summ-dist-feat} (D and E). 

We analyze the number of sentences in the document that achieve the R1, R2, and RL score above a specified threshold \textit{th} with at least one sentence in the summary or spotlight. This approach assumes that if a sentence meets the threshold, its information has been incorporated into the summary or spotlight - i.e. informativeness. However, sentences incorporate information in 
 an abstractive fashion so the threshold \textit{th} needs to be set at a relatively low value.
At a value of $th = 0.3$ our findings of different Rouge (1, 2 and L) scores at the Column `Informativeness' in Table \ref{tab:degree-infor-extract-human} and \ref{tab:degree-infor-extract-gpt4} respectively for human-generated and GPT-4 generated summaries, indicates that a larger number of sentences from the source document are incorporated into summaries compared to spotlights. It signifies that the summary is more informative than the spotlight. 

However, when the threshold value is increased (at a value of $th = 0.8$) — indicating a higher level of extractiveness — the trend reverses, with fewer sentences meeting the threshold in summaries compared to spotlights as shown in `Degree of Extraction' column 
using different Rouge (1, 2 and L) scores 
in Table \ref{tab:degree-infor-extract-human} and \ref{tab:degree-infor-extract-gpt4} respectively for human-generated and GPT-4 generated summaries. The increased extractiveness of the spotlight can also be attributed to its enhanced readability. This trend is consistent across different ROUGE metrics (R1, R2, RL), as shown in the mentioned Tables. 

We have finalized the thresholds after manual verification.  These threshold values are closely similar to the thresholds used by \cite{bhandari-etal-2020-evaluating} for constructing Semantic Content Units (SCUs) as a basis for key fact extraction. The 0.3 threshold serves to identify minimal content overlap—indicating that a generated piece contains at least some of the critical information from the source—while the 0.8 threshold ensures that the content is sufficiently derived from the source, reflecting a high degree of extractiveness.

\begin{table}[]
\begin{adjustbox}{width=\linewidth}
\begin{tabular}{|c|c|cccccc|}
\hline
\multicolumn{2}{|c|}{\multirow{2}{*}{Dataset}} & \multicolumn{3}{|c|}{\textit{threshold=0.3}} &  \multicolumn{3}{|c|}{\textit{threshold=0.8}}                 \\ \cline{3-8} 
                  \multicolumn{2}{|c|}{} & \multicolumn{1}{c|}{R1} & \multicolumn{1}{c|}{R2} & \multicolumn{1}{c|}{RL} &  \multicolumn{1}{c|}{R1} & \multicolumn{1}{c|}{R2} & \multicolumn{1}{c|}{RL} \\\hline
                \multirow{2}{*}{News} & Spot  & \multicolumn{1}{c|}{0.422} & \multicolumn{1}{c|}{0.201} & \multicolumn{1}{l|}{ 0.365} &  \multicolumn{1}{c|}{0.014} & \multicolumn{1}{c|}{0.013} & \multicolumn{1}{c|}{0.014} \\\cline{2-8}
                &  Summ  & \multicolumn{1}{c|}{1.243} & \multicolumn{1}{c|}{0.261} & \multicolumn{1}{l|}{ 0.995}  & \multicolumn{1}{c|}{0.007} & \multicolumn{1}{c|}{0.002} & \multicolumn{1}{c|}{0.007} \\ \hline
                 \multirow{2}{*}{CSPubSum}  &  Spot & \multicolumn{1}{c|}{0.595} & \multicolumn{1}{c|}{0.218} & \multicolumn{1}{l|}{0.513} &  \multicolumn{1}{c|}{0.101} & \multicolumn{1}{c|}{0.020} & \multicolumn{1}{c|}{0.101} \\\cline{2-8}
                 & Summ & \multicolumn{1}{c|}{2.792} & \multicolumn{1}{c|}{1.245} & \multicolumn{1}{l|}{ 2.441} & \multicolumn{1}{c|}{0.056} &  \multicolumn{1}{c|}{0.008} & \multicolumn{1}{c|}{0.047} \\ \hline
                \multirow{2}{*}{Wikipedia}  & Spot  & \multicolumn{1}{c|}{1.519} & \multicolumn{1}{c|}{0.235} & \multicolumn{1}{c|}{0.995} &  \multicolumn{1}{c|}{0.027} & \multicolumn{1}{c|}{0.016} & \multicolumn{1}{c|}{0.026} \\ \cline{2-8}
                &  Summ & \multicolumn{1}{c|}{2.241} & \multicolumn{1}{c|}{0.332} & \multicolumn{1}{c|}{1.204}  & \multicolumn{1}{c|}{0.018} & \multicolumn{1}{c|}{0.010} & \multicolumn{1}{c|}{0.014} \\ \hline
                 Research &  Spot  & \multicolumn{1}{c|}{1.256} & \multicolumn{1}{c|}{0.352} & \multicolumn{1}{c|}{0.765} &  \multicolumn{1}{c|}{0.050} & \multicolumn{1}{c|}{0.021} & \multicolumn{1}{c|}{0.024} \\ \cline{2-8}
                 Presentation & Summ & \multicolumn{1}{c|}{2.856} & \multicolumn{1}{c|}{1.011} & \multicolumn{1}{c|}{1.965}  & \multicolumn{1}{c|}{0.025} & \multicolumn{1}{c|}{0.017} & \multicolumn{1}{c|}{0.016} \\ \hline
\end{tabular}
\end{adjustbox}
\caption{Feature level analysis on Informativeness and Degree of Extraction: spotlight (Spot) vs. GPT-4 generated summary (Summ), Additional Results of Table \ref{tab:qualitative-spot-sum}}
\label{tab:degree-infor-extract-gpt4}
\end{table}

\subsection{Interest Generation aspects of a spotlight} \label{appendix:inquisitiveness_aspect_spot}

\begin{figure*}[!ht]
    \centering
\includegraphics[scale=0.85]{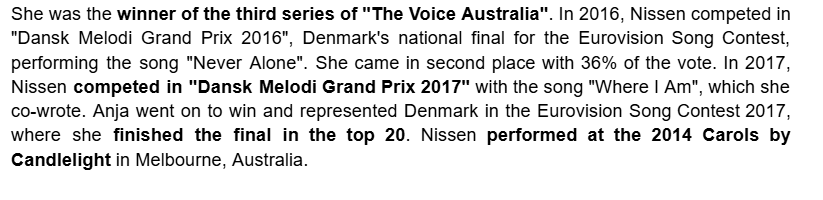}
    \vspace{-8mm}
    \caption{Interest-Generation aspect of a spotlight on Life and Career of Anja Nissen from Wikipedia dataset, where annotators selectively identify the most engaging and narratively significant parts of a text. The \textbf{bold} segments represents the achievements, competitions and performances of Anja Nissen that capture attention and convey intrigue, rather than covering every detail. This mirrors the idea of spotlight generation which focus on a few compelling elements that can spark curiosity and draw readers into the broader narrative.}
    \label{fig:spot_inquisitiveness}
\end{figure*}

The interest generation in spotlights stems from their ability to provide a concise, engaging, and curiosity-inducing preview of the main content. They serve as a hook by emphasizing key points, novel insights, or thought-provoking questions, which draw the audience into exploring the full content. In spotlighting tasks, well-crafted information improves user engagement by offering an informative yet incomplete glimpse, triggering interest to read further. Interest in a spotlight is generated by presenting a novel insight, an unresolved question, or an impactful statement that hints at the significance of the work while leaving room for deep exploration. A well-crafted first sentence, often emphasizing surprise, controversy, or real-world implications, plays a crucial role in capturing attention. We have done an annotation to find out the reasoning behind the interest generation by the same annotators. Figure \ref{fig:spot_inquisitiveness} shows the reasoning behind the generation of interest for a sample from the Wikipedia dataset. A spotlight sparks interest when it introduces new or unconventional ideas, methods, or results in a neutral tone that invites them to explore further. By presenting departures from typical approaches-such as relaxed constraints, adaptive techniques, or novel theoretical insights—it encourages the reader to consider how these changes work and why they matter. Rather than merely summarizing conclusions, the spotlight subtly emphasizes the underlying process, uncertainties, or implications, prompting the reader to mentally engage with open questions or deeper mechanisms, as also seen from Fig \ref{fig:spot_cur1}.

To quantitatively evaluate the effectiveness of spotlights in generating interest, we conducted pairwise comparisons between spotlights and summaries for the same 50 documents. Table \ref{tab:spot_pairwise_results} shows that spotlights were preferred significantly more often than summaries for interest generation. This is evident from the win-rates of spotlight as evident from 3rd Column of Table \ref{tab:spot_pairwise_results} : 68\% for News, 64\% for CSPubSum, 62\% for Wikipedia, and 60\% for Research Presentation which is much higher than summary.

\begin{table*}[htb]
\centering
\begin{adjustbox}{width=0.650\linewidth}
\begin{tabular}{|c|c|c|c|c|}
\hline
\textbf{Dataset} & \textbf{Both of them} & \textbf{Spotlight only} & \textbf{Summary only} & \textbf{None of them} \\
\hline
News & 6 & 34 & 6 & 4 \\\hline
CSPubSum & 11 & 32 & 4 & 3 \\\hline
Wikipedia & 12 & 31 & 1 & 6 \\\hline
Research & \multirow{2}{*}{11} & \multirow{2}{*}{30} & \multirow{2}{*}{1} & \multirow{2}{*}{8} \\
 Presentation&&&&\\
\hline
\end{tabular}
\end{adjustbox}
\caption{Pairwise annotation results showing preferences for spotlight vs. summary across four datasets. ‘Both’ indicates both generates an interest to read the document; ‘None’ indicates neither generated interest.}
\label{tab:spot_pairwise_results}
\end{table*}

\section{Spotlight Generation Process} \label{appendix : kto}

Along with other approaches, we also explore a different preference-based optimization approach - KTO as discussed below.

\noindent \textbf{KTO~\cite{ethayarajh2024kto}:} We initially use the summaries for documents in the dataset to create a binary preference Kahneman-Tversky Optimization (KTO) dataset. Each document from the original dataset contributes one upvoted and one downvoted example. The document paired with the ground truth spotlight forms the upvoted example (indicating a preferred completion), while the document paired with a 
summary forms the downvoted example (denoting undesirable). The rationale behind this is to generate spotlights over summary with higher rewards. This results in the KTO dataset being twice the size of the original dataset, with an equal number of upvoted and downvoted examples. We then apply KTO to the instruction-tuned versions of base models (e.g., Llama-7b-chat-hf, Mistral-7B-Instruct-v0.2) using the KTO dataset, balancing desirable and undesirable weights. Under this setup, we evaluate the performance of the KTO on three LLMs (Mistral-7b, Llama2-7b, and Llama3-8b) by comparing its generations with ground truths on the test set. We also apply KTO on SFT but the results are not better. Among all experiments with different versions of base models, we report the best outcome only (which was achieved when KTO was done directly on the instruct models). These results are denoted in Tables \ref{tab:appendix-news_rouge_gpt}, \ref{tab:appendix-CSPubSum_rouge_gpt}, \ref{tab:appendix-oasum_rouge_gpt}, \ref{tab:appendix-neurips_rouge_gpt} in rows denoted by KTO (Mst-7b), KTO (L2-7b) and KTO (L3-Bb).

\section{Prompts} \label{appendix : prompts}

We use prompting in two stages - data generation, finetune-inference and critique. There are two kinds of prompts - system prompt and user prompt. 

\subsection{Finetune and Inference prompt}

\textbf{system prompt} - You are a helpful, respectful and honest assistant. Your task is to write spotlight of documents. A spotlight is a short, concise overview of the document. It is meant to spark curiosity in the reader to read the entire article. But it does not provide much coverage of the content of the document and that is how it differs from a summary. Below is an instruction, paired with a document that provides further context.\\ 
\textbf{user prompt} - 
\#\#\# Instruction:\\
\#\#\# Document: \{document\}\\
\#\#\# Response: \\ \\
Specific for different datasets are:\\
\textbf{Wikipedia. } Write the spotlight of the following document based on the \{category\} of \{article title\}. The spotlight need not have detailed information coverage but should include only the key points that makes the reader curious about the entire article. \\
\textbf{CSPubSum. } Write the spotlight of the following scientific article entitled \{main\_title\} presented as a document. The spotlight need not have detailed information coverage but should include only the key points that makes the reader curious about the entire article. \\
\textbf{News. } Write a headline for the following news article presented as document. Also include a short description of the article in not more than 4 sentences that can be presented as its highlight. The headline together with the highlight is the spotlight for the news article. It need not have detailed information coverage but should make the reader curious about the news article. \\
\textbf{Research Presentation. } Write the spotlight of the following scientific article presented as a document. The spotlight need not have detailed information coverage but should include only the key points that makes the reader curious about the entire article. \\

\subsection{LLM Critique}

For LLM (GPT-4, Gemini) Critique, we ask the LLM with system and user prompt for evaluating the model-generated spotlights.

\textbf{system}: You are an AI assistant who is to evaluate the spotlight of a textual document. You need to return a score between 0 and 1 reflecting the quality of the generated spotlight based on some criteria. A spotlight is a short summary of some of the most important sentences of a document that creates a curiosity to read the actual document.\\
\textbf{user}: You are given a document and the corresponding spotlight of the document generated by a language model as follows.  \\
Document: \{document\} \\
Ground truth spotlight : \{label spotlight\} \\
spotlight generated: \{model generated spotlight\} \\
Evaluate the above spotlight for the document in terms of each of the following criteria and return only a score between 0 and 1 without any explanation:
\begin{itemize}
    \item Rate the spotlight – whether the spotlight is good or bad. A good spotlight provides a concise, captivating narrative that emphasizes a document’s most intriguing aspects, fostering reader engagement. A bad spotlight either fragments key points without cohesion or overwhelms with excessive detail, failing to engage the reader effectively.
\end{itemize}

\section{Related Work} \label{rel_works}
We study the Spotlight generation task in diverse directions and various concepts similar to the spotlight.

\noindent \textbf{1. Summarization:} Researchers focus on summarization task on different domains - news documents~\cite{du2020news,goyal2022news, huang2014comparative,fikri2021semantic}, scientific articles~\cite{cohan2016revisiting,cohan2017scientific, cachola2020tldr} and other different areas - social media~\cite{mukherjee2022mtlts,chakraborty2019tweet, mullick2024leveraging}, wikipedia~\cite{ta2023wikides,frefel2020summarization} , dialog~\cite{mullick2024long} etc. TLDR~\cite{cachola2020tldr}, TLDR9+~\cite{sotudeh2021tldr9+}, CiteSum~\cite{mao2022citesum}, \cite{narayan2018don} work on extreme summarization of main concepts of articles. People also focus on different summarization - query focused~\cite{vig2022exploring,baumel2016topic,liu2024querysum}, layman summarization~\cite{goldsack2023overview,salaun2022conditional,luo2022readability}, goal oriented~\cite{goldstein2007genre,ham2020end,jin2024comprehensive}, persona-based~\cite{mullick2024persona} etc. CTRLSum~\cite{he2020ctrlsum}, \cite{zhang2023macsum} and ~\cite{ribeiro2023generating} focus on summaries with controllable readability. Researchers also work on summarizing documents with respect to key elements~\cite{ryu2024key,wang2023element}.
But, the problem of spotlight generation differs from these summarization tasks. 

\noindent \textbf{2. Highlight Generation:} People focus on highlight detection task in different directions - story highlight~\cite{woodsend2010automatic}, research paper highlight~\cite{rehman2023research,rehman2023generation}, micro-blog news highlight~\cite{wei2018utilizing}, Teaser generation~\cite{karn2019news}, highlight span extraction~\cite{arumae2019towards,cho2020better}. But, these work mostly focus on extracting multiple spans as a highlight. Our works differ from the fact that we generate most important part of the article which will entice reader to read through the article. 

\noindent \textbf{3. Headline/Title Generation:} Researchers aim at developing different approaches - sequence prediction~\cite{colmenares2015heads} for headline generation, personalized headline generation~\cite{ao2021pens,ao2023put}, self-attention based headline generation~\cite{gavrilov2019self}. But our task is different from traditional headline generation work which is too short. 

\noindent \textbf{4. Teaser Production:} Teasers serve as a powerful strategy for promoting content across various domains. Earlier works cover teaser generation from long documents~\cite{xu2024teasergen}, news article~\cite{karn2019news}, ted-talks~\cite{vico2022ted} etc.


Our work differs from all of the above as Spotlight has different features than summary, highlight, headline and teaser, and serves different perspectives. This concept of spotlight is also different from the notion of entity-extraction~\cite{mullick2024matscire,guha2021matscie,mullick2022using, mullick2021rte}, opinion context detection~\cite{mullick2017generic, mullick2016graphical, mullick2018identifying, mullick2018harnessing, mullick2019d,mullick2017extracting}.
intent phrase identification task in specialized contexts \cite{mullick2022framework, mullick2022fine, mullick2023novel, mullick2023exploring, mullick2022evaluation, mullick2024pointer,mullick2023intent,mullick2024intent}.

\section{Experimental Settings} \label{appendix: experiments}

\subsection{Time and GPU} \label{time-gpu}
We experiment on 80GB A100 GPU with GPU clock cycle 210 MHz. The finetuning and inference time of our finetuned models are in Table \ref{tab:time-gpu}. 

\begin{table}[!ht]
    \centering
    \begin{adjustbox}{width=0.65\linewidth}    
    \begin{tabular}{|c|c|c|}\hline
         Model & Finetune Time & Inference Time\\
         \hline
         L2-7b-FT & 20 hrs & 2hrs 30 mins \\\hline
         DPO (L2-7b) & 40 hrs & 2hrs 30 mins \\\hline
         KTO (L2-7b) & 96 hrs & 2hrs 30 mins \\\hline
         Mst-7b-FT & 17 hrs & 2hrs 10 mins \\\hline
         DPO (Mst-7b) & 36 hrs & 2hrs 10 mins \\\hline
         KTO (Mst-7b) & 83 hrs & 2hrs 10 mins \\\hline
         L3-8b-FT & 22 hrs & 3hrs 20 mins \\\hline
         DPO (L3-8b) & 44 hrs & 3hrs 20 mins \\\hline
         KTO (L3-8b) & 91 hrs & 3hrs 20 mins \\\hline
    \end{tabular}
    \end{adjustbox}
    \caption{Model Training Time [using 80GB A100 GPU]}
    \label{tab:time-gpu}
\end{table}

\subsection{Hyperparameter Optimization} \label{model-config}
For all fine-tuning (SFT) and alignment (DPO, KTO) experiments, we use QLoRA with 4-bit quantization, LoRA \(r=64\), optimization is done with AdamW optimizer, and an effective training batch size of 16 (with gradient accumulation) over 4 epochs. For SFT, we use a learning rate of \(5 \times 10^{-5}\) with linear schedule. For DPO, we use \(\beta=0.01\) and a learning rate of \(5 \times 10^{-6}\) with cosine schedule. For KTO, we use \(\beta=0.01\), with both desirable and undesirable weights set to default 1 and a learning rate of \(5 \times 10^{-6}\) with linear schedule. 

\subsection{Train-validation-test Split}
Table \ref{tab:spotlight_dataset-train-test} shows train-validation-test splits of dataset for experiments.

\begin{table}[!htb]
\vspace{-2mm}
    \centering
    \begin{adjustbox}{width=0.8\columnwidth}
    \begin{tabular}{|c|c|c|c|c|c}
        \hline
         Dataset & Total& Train & Validation & Test \\\hline
        News Category & 14080 & 13000 & 80 & 1000\\\hline
        CSPubSum & 7591 & 6408 & 80 & 1103 \\\hline 
        Wikipedia & 17323 & 14279 & 500 & 2544\\\hline
        Research Presentation & 1230 & 1000 & 100 &130 \\\hline
    \end{tabular}
    \end{adjustbox}
    \vspace{-2mm}
    \caption{Statistics of the spotlight datasets used}
    \label{tab:spotlight_dataset-train-test}
\end{table}

\section{Human Annotations} \label{appendix : human_annotations}


\subsection{\textbf{Task 1 - Identify Interest-Generation}}  \label{prolific_inquisitiveness}

A paragraph describing an article is provided to you and you are asked whether reading that paragraph generates further interests to read the entire document. Use your judgment and perspective to make this determination, ensuring the information is relevant and clear to you. 


\paragraph{Annotation Guidelines for Comparative Rating:} 
We provide instructions with explanations of curiosity generation after reading the provided paragraph (either a spotlight or summary). User is asked to identify if the paragraph generates curiosity to read the entire article as shown in Fig \ref{fig:spot_cur}. 



\paragraph{Instructions:} The instructions are the following - 
In this study, you will assess whether a brief description of an article sparks your interest in reading the entire article. You will be provided with a Short Description (SD) for an article.

Carefully review the Short Description. No additional software download is required. Use a browser, preferably Google Chrome, and ensure a stable internet connection.
Allocate time judiciously for crafting each of the summaries based on the provided 20 SD instances.

After completion, you will be asked to provide feedback on the generation exercise, platform interaction, and details about your academic background, age, country of birth or experience.

\paragraph{Payment Requirements:} Upon completing the study, click on the provided link containing the completion code to redirect you to the Prolific platform. Payment will be processed within one to two weeks.

\paragraph{Ethical Considerations:} Adhere to strict confidentiality and data protection standards to ensure privacy. If you have concerns or questions, feel free to reach out, as this study aligns with ethical guidelines.

This study aims to harness diverse perspectives, including those of academic professionals, to refine the curiosity generation for enhanced utility in various contexts.

Please do not use ChatGPT/GPT4 or any Large Language Models - all responses must be produced by human annotators. It is a strict instruction and will be checked manually - if any issue: the submission will be rejected and re-doing will be required.

\paragraph{
Participant Prescreening criteria:}
\begin{itemize}
    \item Age above 24
    \item Primary Language is English
    \item Minimum Graduation degree is either \textit{Graduate degree (MA/MSc/MPhil/other)} or \textit{Doctorate degree (PhD/other)}
    \item Approval Rate in platform > 85\% and minimum number of previous submissions > 30.
\end{itemize}

\subsection{\textbf{Task 2 - Summary Generation}} \label{prolific_summ_gen}

In this task, you are asked to generate a summary. You will be provided with a section name and a source document (link)  – wikipedia or news article, etc. Read the section of the Source Document and craft summaries. Use your understanding and perspective to tailor the information in a way that is most relevant and comprehensible.

\paragraph{Instructions:} Carefully review the Source Document. No additional software download is required. Use a browser, preferably Google Chrome, and ensure a stable internet connection. Allocate time judiciously for crafting each of the summaries based on the provided 10 SD instances. After completion, you will be asked to provide feedback on the generation exercise, platform interaction, and details about your academic background, age, country of birth, and any medical background or experience with the summary generation task. 
Your summary length (word count) must be around 15\% ($\pm$ 20 words) of the document. For example if the Source Document has 1000 words, the summary must have between 130 to 170 words. Please do not use ChatGPT/GPT4 or any Large Language Models - all reponse should be generated by human properly. It is a strict instruction and will be checked manually - if any issue: it will be rejected and re-doing will be required.

\paragraph{Payment Requirements.} Upon completing the study, click on the provided link containing the completion code to redirect you to the Prolific platform. Payment will be processed within one to two weeks.

\paragraph{Ethical Considerations.} Adhere to strict confidentiality and data protection standards to ensure privacy. If you have concerns or questions, feel free to reach out, as this study aligns with ethical guidelines.

\paragraph{Participant Prescreening criteria.}
\begin{itemize}
    \item Age above 24
    \item Primary Language is English
    \item Minimum Graduation degree is either \textit{Graduate degree (MA/MSc/MPhil/other)} or \textit{Doctorate degree (PhD/other)}
    \item Approval Rate in platform > 85\% and minimum number of previous submissions > 30.
\end{itemize}

\subsection{\textbf{Task 3 - Mini-Story Validation}} \label{prolific_mini_story}

In this study, you are tasked with evaluating paragraphs to determine which one aligns more closely with the characteristics of a mini story. You will be provided with two paragraphs – which highlights some research work. You will be provided with pairs of paragraphs. Your job is to carefully read both paragraphs in each pair and decide which one feels more like a mini story – that is, concise, engaging, and complete with a clear sense of narrative or resolution. Use your understanding and perspective to tailor the information in a way that is most relevant and comprehensible.

\textbf{Annotation Guidelines for Comparative Rating:} 
We provide instructions with explanations of mini-story after reading the provided paragraph (either a spotlight or summary/highlight). User is asked to identify which of the either paragraphs or both the paragraphs are a mini-story.

\textbf{Instructions:} The instructions are the following  

Review both paragraphs in each pair attentively.
Select the paragraph that you believe better captures the essence of a mini story.
No prior knowledge is required, and no additional software download is needed. Use a browser, preferably Google Chrome, and ensure a stable internet connection.
Allocate sufficient time for each comparison to make thoughtful decisions on the provided 10  instances.

After completion, you will be asked to provide feedback on the generation exercise, platform interaction, and details about your academic background, age, country of birth or experience.

\paragraph{Payment Requirements:} Upon completing the study, click on the provided link containing the completion code to redirect you to the Prolific platform. Payment will be processed within one to two weeks.

\paragraph{Ethical Considerations:} Adhere to strict confidentiality and data protection standards to ensure privacy. If you have concerns or questions, feel free to reach out, as this study aligns with ethical guidelines.

Please do not use ChatGPT/GPT4 or any Large Language Models - all reponse should be generated by human properly. It is a strict instruction and will be checked manually - if any issue: the submission will be rejected and re-doing will be required.

Participant Prescreening criteria:-
\begin{itemize}
    \item Age above 24
    \item Primary Language is English
    \item Minimum Graduation degree is either \textit{Graduate degree (MA/MSc/MPhil/other)} or \textit{Doctorate degree (PhD/other)}
    \item Approval Rate in platform > 85\% and minimum number of previous submissions > 30.
\end{itemize}

\begin{table}[!th]
\vspace{-2mm}
    \centering
    \begin{adjustbox}{width=\columnwidth}
    \begin{tabular}{|c|c|c|c|}
        \hline
         Dataset & Both Better(\%) & Spotlight (\%) & Summary (\%) \\\hline
        News Category & 72 & 16 & 12\\\hline
        CSPubSum & 72 & 16 & 8 \\\hline 
        Wikipedia & 68 & 20 & 12\\\hline
        Research Presentation & 80 & 12 & 8 \\\hline
    \end{tabular}
    \end{adjustbox}
    \vspace{-2mm}
    \caption{Mini-Story Validation Study - Prolific}
    \label{tab:appendix_mini_story_prolific}
\end{table}

\color{black}

\section{Examples}  \label{appendix: examples}

\begin{figure*}[!ht]
    \centering
    \includegraphics[scale=0.9]{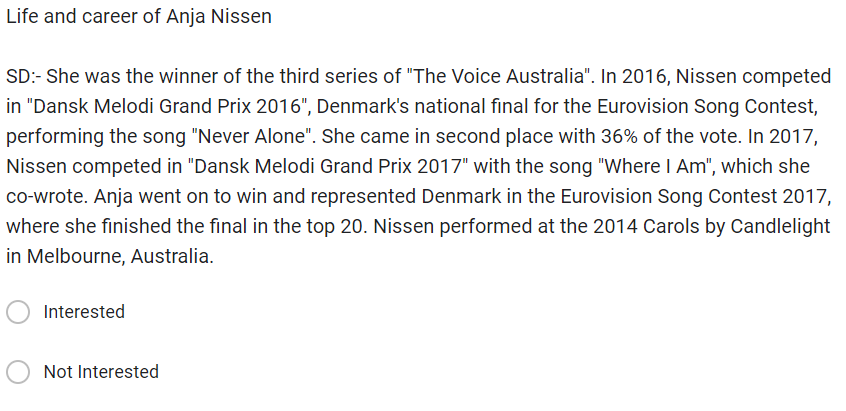}
    \caption{Spotlight curiosity experiment example snapshot}
    \label{fig:spot_cur}
\end{figure*}

One example of the human annotation interface is shown in Fig \ref{fig:spot_cur}.

\section{Experimental Results}

We analyzed the performance of traditional and LLM-based approaches, including the preference optimization techniques using various referenceless metrics and LLM-critiquing. We also assess different variations of DPO with held-out data.

\subsection{Performance of Traditional Baseline Approaches}

In addition to the results in Table \ref{tab:cspubsum_rouge_gpt}, we have also explored different traditional approaches  for spotlight generation and results for News, CSPubSum, Wikipedia and Research Presentation datasets are shown in Table \ref{tab:news_rouge_gpt_baseline}, \ref{tab:cspubsum_rouge_gpt_baseline}, \ref{tab:oasum-domain_rouge_gpt_baseline} and \ref{tab:neurips_rouge_gpt_baseline}. It is seen that even simple SFT-based approaches on LLMs perform much better than these traditional approaches across all metrics. This performance gap may be due to the advanced architectures and targeted optimizations of LLM models which are fine-tuned specifically for task performance and user preferences, whereas the older models do not benefit from the same level of refinement and cutting-edge techniques, leading to their comparatively lower performance. 

\subsection{Performance of DPO vis-a-vis other LLM-based approaches} \label{appendix:dpo_comparison_and_discussion} 

In addition to the results presented in Table \ref{tab:rouge_gpt}, we demonstrate the effectiveness of the DPO and KTO approach for spotlight generation across various evaluation criteria for Llama3-8b (L3-8b), Llama2-7b (L2-7b) and Mistral-7b (Mst-7b) large language model (LLM) for each datasets in Table \ref{tab:appendix-news_rouge_gpt} (for News), \ref{tab:appendix-CSPubSum_rouge_gpt} (for CSPubSum), \ref{tab:appendix-oasum_rouge_gpt} (for Wikipedia) and \ref{tab:appendix-neurips_rouge_gpt} (for Research Presentation). We report evaluations for Llama2-7b, Mistral-7b and Llama3-8b finetuning {abbreviated as L2-7b-FT, Mst-7b-FT and L3-8b-FT}, general instruction tuned vanilla model {L2-7b-VA, Mst-7b-VA and L3-8b-VA}, DPO {DPO(L2-7b), DPO(Mst-7b) and DPO(L3-8b)} and KTO {KTO(L2-7b), KTO(Mst-7b) and KTO(L3-8b)}. It is seen that

\begin{itemize}
    \item DPO consistently outperforms vanilla models, supervised fine-tuning (SFT), KTO and GPT-4 (1-shot) (in most of the cases) proving its superiority in spotlight generation.
    \item SFT only approach is a close second, however, DPO consistently improves over SFT methods  indicating its utility and  the potency of SFT when the reference negative examples (here summaries) are not available
    \item Vanilla L3-8b-VA performs the worst, struggling in both traditional and advanced evaluations, highlighting the importance of fine-tuning or DPO across all datasets.
    \item for \textit{News, CSPubSum and Research Presentation} datasets, Llama3-8b fine-tuning and DPO, both perform better than GPT-4 (1-shot) for traditional metrics whereas for \textit{Research Presentation}, GPT-4 (1-shot) performs better than the fine-tuning model but worse than DPO.
    \item GPT-4 (1-shot) is strong in UniEval Relevance and Critique (it is best among all in \textit{Wikipedia}) but falls behind in ROUGE and BERTScore, suggesting it produces more readable but not necessarily more extractive spotlights.
    \item For Wikipedia data, all models show higher Rouge scores with lower UniEval relevance/GPT-4 Critique suggests that while the spotlights have significant lexical overlap with the reference, they may lack deeper semantic coherence.
    \item Notably, the Llama3-8b based DPO model achieves statistically significant improvements (p < 0.05) over other model variations. 
    \item Across all the datasets, both Llama2 and Mistral models consistently show that DPO optimization leads to superior performance over KTO—evidenced by higher ROUGE, BERT‑F1, UniEval relevance, and GPT‑4 Critique scores. This aggregated trend reinforces the robustness of the DPO approach in spotlight generation across diverse domains.
    \item  The experimental results indicate that the DPO-based approach outperforms large LLMs (GPT-3.5, Gemini and GPT-4) in 1‑shot evaluations for spotlight generation.
    \item Also, to validate the GPT-4 critique scores against robust human judgment, we conducted an additional human annotation study similar to \cite{mullick2024persona} to evaluate the alignment between human judgment and automated assessments for 20 documents for each dataset. In this study, annotators were presented with the model-generated spotlight and asked to categorize them into one of two options - `good' or `bad'. Then, we calculate the Point-biserial correlation coefficient between GPT-4 critique scores and human judgment. We obtain an overall average of 0.751 point-biserial correlation coefficient between them, which validates GPT-4 critique scores are robust and reliable against human judgment studies.
\end{itemize} 

\subsection{Reference-less metrics for Spotlight Generation}

In Table \ref{tab:appendix_rouge_gpt_evals_ref_free}, we focused on the best-performing models—specifically, the DPO-based Llama3-8b and GPT-4 (1-shot) approaches—and assessed their outputs using reference-less metrics, including UniEval (Coherence, Consistency, Fluency) and G-Eval(Relevance, Coherence, Factuality, Conciseness). The results indicate that while GPT-4 (1-shot) generates spotlights that are notably readable and contextually relevant, the DPO-based model consistently excels in semantic coherence and consistency across diverse datasets. These findings suggest that the targeted fine-tuning strategy inherent in the DPO approach effectively enhances the model’s capacity to capture the nuanced characteristics of high-quality spotlights, underscoring its robustness for applications demanding high semantic fidelity.

\subsection{Different Variations of DPO with Held-out Data} 
\label{app:dpo-variations}
Table \ref{tab:rouge_gpt}
demonstrates that DPO applied to the larger, fine-tuned Llama3-8b model consistently outperforms other LLM-DPO combinations across all datasets and metrics.
This suggests that model size and fine-tuning play a crucial role in DPO's effectiveness. We have explored different DPO configurations using data and fine-tuning as in Table \ref{tab:dpo-three-sets}:

\begin{itemize}
    \item \textit{DPO (vnl)}: all data for DPO directly on vanilla (pre-trained) LLMs without fine-tuning. 
    \item \textit{DPO (n/2)}: 50\% data for the supervised fine-tuning model and rest 50\% data for DPO 
    \item \textit{DPO (all)}: all data for fine-tuning followed by all data for DPO. 
\end{itemize} 
Interestingly, option 2 (50\% data each for fine-tuning and DPO) achieves the best results for Llama3-8b based DPO variants across Rouge-1 (R1), Rouge-2 (R2), Rouge-L (RL), BERTScore F1 (BeF1), UniEval Relevance (Rel) and GPT-4 critique scores - 
Rating (Rat) for all four datasets.
Using 50\% of the data for supervised fine-tuning allows the model to learn directly from high-quality examples of spotlights, establishing a strong baseline understanding of the desired output style. The remaining 50\% is then used for DPO, where the model refines its decision-making by comparing preferred spotlight against summary. This two-stage approach combines the strengths of learning from examples with contrastive learning, ensuring the model not only understands what a good spotlight looks like but also how to distinguish it from a summary.
This suggests a potential sweet spot between model capacity and fine-tuning for optimal DPO performance shown in Table \ref{tab:dpo-three-sets}. 


\subsection{Robustness of the GPT-4 Critiquing} 

To validate our evaluation framework, we access the robustness of the GPT-4 critiquing system. We evaluate all the DPO model generated spotlights in the test set with Gemini-Pro model keeping the same prompts and criteria as used earlier for GPT-4. Gemini scores are described in Table \ref{tab:gemini-dpo} and values are aligned with GPT-4 scores. The GPT-4-Gemini correlation coefficient is 0.98 showing perfect alignment. This verifies that the GPT-4 based evaluation is impartial and robust.

\subsection{Cross-Domain  Generalization of DPO} 

Our primary focus is on establishing a solid foundation for spotlight generation and demonstrating the efficacy of our proposed method within the chosen domains. Table \ref{tab:rouge_gpt} shows a  considerable performance improvement of models when trained using SFT or DPO on a specific domain, but a question remained on the generalizability capability of the proposed models. We take the best-performing model (Llama3-8b DPO model) trained on one dataset and test it on the other dataset. An additional set of experiments in Table \ref{tab:out-of-domain} shows the generalizability of the best performing model, DPO (L3-8b) from Table \ref{tab:rouge_gpt}.  indicating that the DPO‑enhanced model robustly generalizes to unseen domains without significant degradation. This outcome reinforces the model’s consistency and effectiveness across varied datasets. The results show comparable outcome to in-domain dataset assessment as shown in Table \ref{tab:rouge_gpt}.

\begin{table*}[]
    \centering
    \begin{adjustbox}{width=0.95\linewidth}
    \begin{tabular}{|c|c|c|c|c|c|c|c|}
        \hline
        \multirow{2}{*}{\textbf{Model}} & \multicolumn{3}{|c|}{\textbf{UniEval}}& \multicolumn{4}{|c|}{\textbf{G-Eval}}\\
        \cline{2-8}
        & \textbf{Coherence} & \textbf{Consistency} & \textbf{Fluency} &  \textbf{Relevance} &  \textbf{Coherence} & \textbf{Factuality} &  \textbf{Conciseness}\\
        \hline
\multicolumn{8}{|c|}{News} \\\hline
L3-8b-VA& 75.3 & 85.0 & 80.1 & 68.6 & 73.2 & 75.4 & 72.5\\\hline
GPT-4 (1-shot) & \textbf{83.8} & 90.2 & 88.2 & 82.2 & \textbf{85.8} & \textbf{88.5} & 83.4\\\hline
L3-8b-FT& 82.8 & 93.0 & 89.9 & 81.5 & 84.4 & 87.1 & 83.1 \\\hline
DPO (L3-8b)& 83.4 & \textbf{94.8} & \textbf{90.1} & \textbf{83.6} & 85.2 & 88.4 & \textbf{84.4}\\\hline
 \multicolumn{8}{|c|}{CSPubSum}\\\hline
L3-8b-VA&  62.5 & 70.5 & 72.3 & 60.3 & 63.4 & 64.4 & 62.8\\\hline
GPT-4 (1-shot) & 66.2 & \textbf{77.8} & 87.2 & 66.9 & 69.1 & 72.5 & 68.2\\\hline
L3-8b-FT & \textbf{69.8} & 74.4 & 88.3 & 64.8 & 68.6 & 70.1 & 67.4\\\hline
DPO (L3-8b)&  69.5 & 75.3 & \textbf{89.7} & \textbf{67.4} & \textbf{70.6} & \textbf{73.0} & \textbf{69.5}\\\hline
\multicolumn{8}{|c|}{Wikipedia}\\\hline 
L3-8b-VA & 65.2 & 73.8 & 80.2 & 62.1 & 64.5 & 66.7 & 63.5 \\\hline
GPT-4 (1-shot) & 73.1 & \textbf{81.9} & 88.6 & \textbf{71.3} & \textbf{75.5} & \textbf{79.1} & \textbf{73.2}\\\hline
L3-8b-FT& 73.7 & 80.4 & 90.6 & 70.3 & 74.2 & 78.7 & 72.8\\\hline
DPO (L3-8b)& \textbf{75.9} & 79.2 & \textbf{91.2} & 69.0 & 73.6 & 77.5 & 71.2\\\hline
 \multicolumn{8}{|c|}{Research Presentation} \\\hline
L3-8b-VA & 80.4 & 69.6 & 85.3 & 69.2 & 79.1 & 73.4 & 70.5 \\\hline
GPT-4 (1-shot) & \textbf{88.2} & \textbf{78.9} & 86.3 & 74.3 & 79.4 & 78.5 & 75.5\\\hline
L3-8b-FT & 86.9 & 72.3 & 88.3 & 75.8 & 80.2 & 78.1 & 76.4\\\hline
DPO (L3-8b) &  84.8 & 76.6 & \textbf{89.7} & \textbf{77.4} & \textbf{81.9} & \textbf{80.2} & \textbf{77.5}\\\hline
    \end{tabular}
    \end{adjustbox}
    \vspace{-2mm}
    \caption{UniEval and G-Eval  reference-free evaluation criteria Results (The best outcomes are in \textbf{Bold}) for each dataset, Additional Results of Table \ref{tab:rouge_gpt}
    }
    \label{tab:appendix_rouge_gpt_evals_ref_free}
\end{table*}

\begin{table*}[!htb]
    \centering
    \begin{adjustbox}{width=0.65\linewidth}
    \begin{tabular}{|c|c|c|c|c|c|c|c|c|}
        \hline
        \multirow{2}{*}{\textbf{Model}} &  \multicolumn{6}{|c|}{\textbf{Traditional Metrics}} &  \multicolumn{1}{|c|}{\textbf{UniEval}} &  \multicolumn{1}{|c|}{\textbf{GPT-4}}\\
        \cline{2-8}
          & \textbf{R1} & \textbf{R2} & \textbf{RL}& \textbf{Mt} & \textbf{Bl} & \textbf{BeF1} & Rel & \textbf{Critique}\\
        \hline
L2-7b-FT& 34.9 & 14.6 & 27.9 & 31.2 & 5.5 & 81.0 & 80.6 & \textbf{89.6}\\\hline
Mst-7b-FT & 32.1 & 12.3 & 25.0 & 28.3 & 3.8 & 80.4 & 79.9 & 89.4\\\hline
L3-8b-FT& \textbf{36.5} & \textbf{15.8} & \textbf{28.1} & \textbf{32.8} & \textbf{6.5} & \textbf{81.5} & \textbf{82.2} & 89.4\\\hline\hline
LF & 14.9 & 6.1 & 10.9 & 22.2 & 1.9 & 71.9 &75.7 & 48.8\\\hline
T5-FT & 18.0 & 10.6 & 18.0 & 25.0 & 4.0 & 76.0 & 77.5 & 69.2\\\hline
BART & 19.6 & 8.3 & 14.2 & 28.4 & 3.1 & 75.8 & 81.9 & 71.2\\\hline
Pegasus	& 20.1 & 7.1 & 14.4 & 24.7 & 2.4 & 75.7 & 77.2 & 70.7\\\hline
Falcon & 15.0 & 2.7 & 10.0 & 18.0 & 0.3 & 72.0 & 76.3 & 48.0\\\hline
TLDR & 29.9 & 11.1 & 23.1 & 24.0 & 1.5 & 79.0 & 77.3 & 69.3\\\hline
    \end{tabular}
    \end{adjustbox}
    \caption{Traditional, UniEval and GPT-4 Critique evaluations on News Dataset - comparison of baselines with SFT, Additional Results of Table \ref{tab:cspubsum_rouge_gpt}}
    \vspace{-1mm}
    \label{tab:news_rouge_gpt_baseline}
\end{table*}

\begin{table*}[!htb]
    \centering
    \begin{adjustbox}{width=0.75\linewidth}
    \begin{tabular}{|c|c|c|c|c|c|c|c|c|}
        \hline
         \multirow{2}{*}{\textbf{Model}} &  \multicolumn{6}{|c|}{\textbf{Traditional Metrics}} &  \multicolumn{1}{|c|}{\textbf{UniEval}} &  \multicolumn{1}{|c|}{\textbf{GPT-4}}\\
        \cline{2-8}
          & \textbf{R1} & \textbf{R2} & \textbf{RL}& \textbf{Mt} & \textbf{Bl} & \textbf{BeF1} & \textbf{Rel} & \textbf{Critique}\\
        \hline
 L2-7b-VA& 15.5 & 3.8 & 11.4 & 17.2 & 0.5 & 70.6 & 62.3 & 72.6\\\hline
 L2-7b-FT& 34.9 & 14.6 & 27.9 & 31.2 & 5.5 & 81.0 & 80.6 & 89.6\\\hline
 DPO (L2-7b) & 35.5 & 15.2 & 28.5 & 31.6 & 6.1 & 81.2 & 82.3 & 89.7\\\hline
 KTO (L2-7b)& 31.3 & 11.9 & 23.6 & 24.3 & 2.8 & 79.5 & 81.9 &  82.3 \\\hline 
Mst-7b-VA& 20.1 & 5.6 & 14.2 & 22.9 & 0.7 & 74.2 & 70.1 & 76.5\\\hline
Mst-7b-FT & 32.1 & 12.3 & 25.0 & 28.3 & 3.8 & 80.4 & 79.9 & 89.4\\\hline
DPO (Mst-7b)& 34.4 & 12.9 & 25.8 & 29.8 & 4.3 & 82.5 & 82.6 & \textbf{91.4}\\\hline
KTO (Mst-7b)& 26.3 & 9.5 & 19.7 & 24.2 & 2.1 & 70.6 & 80.3 & 89.3\\\hline
L3-8b-VA& 15.9 & 6.8 & 11.8 & 25.2 & 2.0 & 73.8 & 70.6 & 78.0\\\hline
 L3-8b-FT& 36.5 & 15.8 & 28.1 & 32.8 & 6.5 & 81.5 &  82.2 & 89.4\\\hline
DPO (L3-8b)& \textbf{36.8} & \textbf{16.0} & \textbf{28.6} & \textbf{33.5} & \textbf{6.9} & \textbf{83.6} & \textbf{84.7}  & 91.3\\\hline
KTO (L3-8b) & 28.8 & 11.1 & 21.7 & 28.9 & 2.7 & 77.9 & 81.4   & 89.3\\\hline
GPT-3.5 (1-shot) & 28.3 & 11.3 & 22.0 & 23.7 & 2.3 & 77.1 & 82.5  & 80.4\\\hline
GPT-4 (1-shot) & 31.3 & 12.1 & 22.3 & 32.8 & 3.4 & 79.5 &  83.1 & 90.1\\\hline
Gemini (1-shot) & 29.3 & 14.1 & 23.4 & 23.9 & 3.4 & 80.6 & 82.7 & 90.0\\\hline
    \end{tabular}
    \end{adjustbox}
    \vspace{-2mm}
    \caption{Traditional, UniEval and GPT-4 Critique evaluations on News Dataset, (The best outcomes are marked in \textbf{Bold}) Additional Results of Table \ref{tab:rouge_gpt}}
    \label{tab:appendix-news_rouge_gpt}
\end{table*}

\begin{table*}[htb]
    \centering
    \begin{adjustbox}{width=0.65\linewidth}
    \begin{tabular}{|c|c|c|c|c|c|c|c|c|}
        \hline
        \multirow{2}{*}{\textbf{Model}} &  \multicolumn{6}{|c|}{\textbf{Traditional Metrics}} &  \multicolumn{1}{|c|}{\textbf{UniEval}} &  \multicolumn{1}{|c|}{\textbf{GPT-4}}\\
        \cline{2-8}
          & \textbf{R1} & \textbf{R2} & \textbf{RL}& \textbf{Mt} & \textbf{Bl} & \textbf{BeF1} & \textbf{Rel} & \textbf{Critique}\\
        \hline
L2-7b-FT & 28.1 & 8.6 & 20.4 & \textbf{24.8} & 2.7 & 81.3 & 66.6 & 77.8\\\hline
Mst-7b-FT & 25.7 & 6.9 & 17.7 & 17.6 & 1.7 & 77.4 & 65.1 & 77.0\\\hline
L3-8b-FT & \textbf{31.9} & \textbf{9.7} & \textbf{22.6} & 23.6 & \textbf{3.1} & 69.0 & \textbf{67.1} & \textbf{79.5}\\\hline\hline
LF & 16.0 & 4.8 & 11.5 & 16.7 & 1.7 & 67.4 &55.2 &  68.7\\\hline
T5-FT & 20.0 & 5.0 & 16.0 & 15.0 & 2.0 & 71.0 & 63.2 & 75.9\\\hline
BART & 19.1 & 3.3 & 8.3 & 19.3 & 2.4 & 71.9 &64.7&  70.2\\\hline
Pegasus	& 17.8 & 4.1 & 9.2 & 16.9 & 1.7 & 70.8 & 62.5& 60.9\\\hline
Falcon& 17.0 & 3.0 & 11.0 & 15.0 & 0.8 & 74.0 & 55.2 & 38.4\\\hline
TLDR & 19.5  & 5.6 & 14.8 & 11.6 & 0.6 & 77.6 & 62.9 & 72.3\\\hline
    \end{tabular}
    \end{adjustbox}
    \caption{Traditional, UniEval and GPT-4 Critique evaluations on CSPubSum Dataset - comparison of baselines with SFT, Additional Results of Table \ref{tab:cspubsum_rouge_gpt}}
    \label{tab:cspubsum_rouge_gpt_baseline}
\end{table*}

\begin{table*}[htb]
    \centering
    \begin{adjustbox}{width=0.75\linewidth}
    \begin{tabular}{|c|c|c|c|c|c|c|c|c|c|c|}
        \hline
         \multirow{2}{*}{\textbf{Model}} &  \multicolumn{6}{|c|}{\textbf{Traditional Metrics}} &  \multicolumn{1}{|c|}{\textbf{UniEval}} &  \multicolumn{1}{|c|}{\textbf{GPT-4}}\\
        \cline{2-8}
         & \textbf{R1} & \textbf{R2} & \textbf{RL}& \textbf{Mt} & \textbf{Bl} & \textbf{BeF1} & \textbf{Rel} & \textbf{Critique}\\
        \hline
L2-7b-VA& 14.3 & 3.3 & 9.7 & 10.9 & 0.9 & 45.9 & 58.6 & 79.0\\\hline
L2-7b-FT & 28.1 & 8.6 & 20.4 & 24.8 & 2.7 & 81.3 & 64.8 & 77.8\\\hline
DPO (L2-7b) &  30.4 & 9.2 & \textbf{26.8} & 25.1 & 2.8 & 81.6 & 65.1 & 83.1\\\hline
KTO (L2-7b)& 27.7 & 8.1 & 19.0 & 22.9 & 2.5 & 79.7 & 64.6 & 78.6\\\hline
Mst-7b-VA & 13.8 & 2.9 & 9.1 & 10.6 & 0.7 & 46.4 & 59.2 & 79.7\\\hline
Mst-7b-FT & 25.7 & 6.9 & 17.7 & 17.6 & 1.7 & 77.4 & 64.0 & 77.0\\\hline
DPO (Mst-7b) & 29.9 & 7.6 & 20.1 & \textbf{25.8} & 2.9 & 81.4 & 65.4 & 85.0\\\hline
KTO (Mst-7b) & 27.9 & 7.4 & 18.2 & 25.1 & 2.6 & 80.0 &  
 64.8 & 78.1\\\hline
L3-8b-VA& 15.5 & 3.8 & 10.2 & 12.8 & 1.0 & 47.2 & 60.1 & 73.6\\\hline
L3-8b-FT & 31.9 & 9.7 & 22.6 & 23.6 & 3.1 & 81.8 & 67.1 & 79.5\\\hline
DPO (L3-8b)& \textbf{32.3} & \textbf{9.9} & 23.0 & 24.9 & \textbf{3.5} & \textbf{82.3} & \textbf{67.8} & \textbf{89.0}\\\hline
KTO (L3-8b)& 29.1 & 5.4 & 13.2 & 15.1 & 1.6 & 75.0 & 64.6 & 81.0\\\hline
GPT-3.5 (1-shot) & 27.5 & 6.9 & 17.9 & 19.9 & 2.6 & 79.5 & 65.2 & 76.8 \\\hline
GPT-4 (1-shot) & 28.7 & 7.4 & 17.9 & 23.3 & 3.2 & 80.8 & 66.8 & 82.1\\\hline
Gemini (1-shot) & 26.6 & 5.7 & 17.2 & 18.0 & 1.6 & 78.5 & 67.1 & 71.5\\\hline
    \end{tabular}
    \end{adjustbox}
    \vspace{-2mm}
    \caption{Traditional, UniEval and GPT-4 Critique evaluations on CSPubSum Dataset, (The best outcomes are marked in \textbf{Bold}) Additional Results of Table \ref{tab:rouge_gpt}
    }
    \label{tab:appendix-CSPubSum_rouge_gpt}
\end{table*}

\begin{table*}[!ht]
    \centering
    \begin{adjustbox}{width=0.65\linewidth}
    \begin{tabular}{|c|c|c|c|c|c|c|c|c|}
        \hline
        \multirow{2}{*}{\textbf{Model}} &  \multicolumn{6}{|c|}{\textbf{Traditional Metrics}} &  \multicolumn{1}{|c|}{\textbf{UniEval}} &  \multicolumn{1}{|c|}{\textbf{GPT-4}}\\
        \cline{2-8}
          & \textbf{R1} & \textbf{R2} & \textbf{RL}& \textbf{Mt} & \textbf{Bl} & \textbf{BeF1} & \textbf{Rel} & \textbf{Critique}\\
        \hline
L2-7b-FT& 29.3 & 13.6 & 26.2 & 22.8 & 5.9 & 70.0 & 67.2 & 51.4 \\\hline
Mst-7b-FT & 25.5 & 14.5 & 21.8 & 20.3 & 8.4 & 78.8 & 66.9 & 52.8\\\hline
L3-8b-FT& \textbf{42.4} & \textbf{26.0} & \textbf{38.2} & \textbf{38.2} & \textbf{16.6} & \textbf{81.3} & \textbf{69.2} & \textbf{64.8}\\\hline\hline
LF & 16.2 & 4.6 & 12.0 & 20.3 & 1.0 & 70.0 & 63.2 & 43.5\\\hline
T5-FT	& 25.2 & 8.2 & 21.1 & 19.5 & 7.1 & 68.7 & 65.2 & 54.6 \\\hline
BART &	23.9 & 8.5  & 17.6 & 27.5 & 3.3 & 74.6 & 69.2 & 41.3 \\\hline
Pegasus	& 19.8 & 5.5  & 14.2 & 21.9 & 1.9 & 71.9 & 64.2 & 47.8 \\\hline
Falcon & 17.0 & 5.0 & 12.0 & 22.0 & 1.2 & 71.0 & 63.4 & 45.8\\\hline
TLDR & 25.8 & 9.8 & 21.4 & 20.3 & 3.14 & 74.4 & 63.9 & 61.0\\\hline
    \end{tabular}
    \end{adjustbox}
    \caption{Traditional, UniEval and GPT-4 Critique evaluations on Wikipedia Dataset - comparison of baselines with SFT, Additional Results of Table \ref{tab:cspubsum_rouge_gpt}}
    \label{tab:oasum-domain_rouge_gpt_baseline}
\end{table*}

\begin{table*}[]
    \centering
    \begin{adjustbox}{width=0.75\linewidth}
    \begin{tabular}{|c|c|c|c|c|c|c|c|c|}
        \hline
         \multirow{2}{*}{\textbf{Model}} &  \multicolumn{6}{|c|}{\textbf{Traditional Metrics}} &  \multicolumn{1}{|c|}{\textbf{UniEval}} &  \multicolumn{1}{|c|}{\textbf{GPT-4}}\\
        \cline{2-8}
          & \textbf{R1} & \textbf{R2} & \textbf{RL}& \textbf{Mt} & \textbf{Bl} & \textbf{BeF1} & \textbf{Rel} & \textbf{Critique} \\
        \hline
L2-7b-VA& 17.0 & 4.1 & 12.7 & 18.3 & 1.2 & 67.3 & 62.1 & 42.4 \\\hline
L2-7b-FT& 29.3 & 13.6 & 26.2 & 22.8 & 5.9 & 70.0 & 67.2 & 51.4 \\\hline
DPO (L2-7b) & 38.6 & 23.5 & 35.5 & 30.9 & 15.7 & 80.1 & 67.5 & 55.6\\\hline
KTO (L2-7b)& 31.5 & 18.6 & 27.6 & 29.8 & 13.4 & 81.1 & 66.1 & 55.6\\\hline
Mst-7b-VA & 18.9 & 4.9 & 13.7 & 20.4 & 1.4 & 70.2 & 63.4 & 36.4\\\hline
Mst-7b-FT & 25.5 & 14.5 & 21.8 & 20.3 & 8.4 & 78.8 & 66.9 & 52.8\\\hline
DPO (Mst-7b)& 37.2 & 20.8 & 28.1 & 27.1 & 13.9 & 82.9 & 67.1 & 65.2\\\hline
KTO (Mst-7b)& 34.1 & 19.9 & 22.7 & 23.1 & 11.7 & 78.6 & 66.2 & 63.0\\\hline
L3-8b-VA & 20.3 & 6.5 & 15.0 & 23.6 & 2.2 & 71.7 & 65.7 & 52.1\\\hline
L3-8b-FT& 42.4 & 26.0 & 38.2 & 38.2 & 16.6 & 81.3 & 69.2 & 64.8\\\hline
DPO (L3-8b)& \textbf{42.5} & \textbf{26.3} & \textbf{38.4} & \textbf{38.7} & \textbf{16.7} & \textbf{83.4} & 67.3 & 67.3\\\hline
KTO (L3-8b)& 32.0 & 14.9 & 22.4 & 29.1 & 11.5 & 82.0 & 65.2 & 60.0 \\\hline
GPT-3.5 (1-shot) & 29.8 & 11.5 & 23.6 & 22.3 & 4.2 & 77.7 & 68.3 & 60.2 \\\hline
GPT-4 (1-shot) & 34.6 & 13.8 & 26.1 & 30.8 & 7.1 & 80.4 & \textbf{70.2} & \textbf{73.5}\\\hline
Gemini (1-shot) & 29.7 & 15.3 & 26.1 & 28.9 & 5.1 & 79.0 & 68.0 & 61.0\\\hline
    \end{tabular}
    \end{adjustbox}
    \caption{Traditional, UniEval and GPT-4 Critique evaluations on Wikipedia Dataset, (The best outcomes are marked in \textbf{Bold}) Additional Results of Table \ref{tab:rouge_gpt}}
    \vspace{-2mm}
    \label{tab:appendix-oasum_rouge_gpt}
\end{table*}

\begin{table*}[htb]
    \centering
    \begin{adjustbox}{width=0.65\linewidth}
    \begin{tabular}{|c|c|c|c|c|c|c|c|c|}
        \hline
        \multirow{2}{*}{\textbf{Model}} &  \multicolumn{6}{|c|}{\textbf{Traditional Metrics}} &  \multicolumn{1}{|c|}{\textbf{UniEval}} &  \multicolumn{1}{|c|}{\textbf{GPT-4}}\\
        \cline{2-8}
          & \textbf{R1} & \textbf{R2} & \textbf{RL}& \textbf{Mt} & \textbf{Bl} & \textbf{BeF1} & \textbf{Rel} & \textbf{ Critique}\\
        \hline
L2-7b-FT& 29.9 & 7.1 & 16.5 & 14.9 & 2.1 & 77.2 & 74.7 & 80.1\\\hline
Mst-7b-FT& \textbf{36.6} & \textbf{8.9} & \textbf{17.0} & \textbf{18.1} & \textbf{2.2} & \textbf{79.2} & 75.2 & 80.5\\\hline
L3-8b-FT& 28.3 & 6.9 & 14.7 & 13.7 & 1.6 & 76.5 & \textbf{76.7} & \textbf{81.2}\\\hline\hline
LF & 13.2 & 2.2 & 5.5 & 7.9 & 0.5 & 61.5 & 69.7& 68.7\\\hline
T5-FT & 18.3 & 2.3 & 9.6 & 6.6 & 1.1 & 67.0 & 70.7 & 71.4\\\hline
BART & 16.8 & 1.5 & 3.2 & 11.4 & 1.5 & 68.9 & 73.7 &  61.0\\\hline
Pegasus	& 15.5 & 1.1 & 3.5 & 10.2 & 1.2 & 68.8 & 69.7 & 52.9\\\hline
Falcon& 15.2 & 1.6 & 6.2 & 7.0 & 0.6 & 71.0 & 68.7 & 39.6\\\hline
TLDR & 17.1  & 2.1 & 8.9 & 8.6 & 0.4 & 73.6 & 69.7& 67.3\\\hline
    \end{tabular}
    \end{adjustbox}
    \caption{Traditional, UniEval and GPT-4 Critique evaluations on Research Presentation Dataset - comparison of baselines with SFT, (The best outcomes are marked in \textbf{Bold}) Additional Results of Table \ref{tab:cspubsum_rouge_gpt}}
    \vspace{-1mm}
    \label{tab:neurips_rouge_gpt_baseline}
\end{table*}

\begin{table*}[htb]
\vspace{-6mm}
    \centering
    \begin{adjustbox}{width=0.75\linewidth}
    \begin{tabular}{|c|c|c|c|c|c|c|c|c|}
        \hline
         \multirow{2}{*}{\textbf{Model}} &  \multicolumn{6}{|c|}{\textbf{Traditional Metrics}} &  \multicolumn{1}{|c|}{\textbf{UniEval}} &  \multicolumn{1}{|c|}{\textbf{GPT-4}}\\
        \cline{2-8}
          & \textbf{R1} & \textbf{R2} & \textbf{RL}& \textbf{Mt} & \textbf{Bl} & \textbf{BeF1} & \textbf{Rel} & \textbf{Critiquie}\\
        \hline
L2-7b-VA& 12.5 & 3.1 & 9.6 & 8.1 & 0.7 & 46.3 & 68.5 & 78.5\\\hline
L2-7b-FT& 29.9 & 7.1 & 16.5 & 14.9 & 2.1 & 77.2 & 74.7 & 80.1\\\hline
DPO (L2-7b)& 30.5 & 6.9 & 16.6 & 15.4 & 1.9 & 77.5 & 75.1 & 85.2\\\hline
KTO (L2-7b)& 28.5 & 6.6 & 14.9 & 14.5 & 2.0 & 75.3 & 74.4 & 83.6\\\hline 
Mst-7b-VA& 13.6 & 2.6 & 9.0 & 10.2 & 0.6 & 47.2 & 69.3 & 79.7\\\hline
Mst-7b-FT& 36.6 & 8.9 & 17.0 & 18.1 & 2.2 & 79.2 & 75.2 & 80.5\\\hline
DPO (Mst-7b)& 37.3 & 8.8 & 17.5 & 18.6 & 2.6 & 79.3 & 75.8 & 85.6\\\hline
KTO (Mst-7b)& 34.8 & 7.3 & 16.5 & 18.0 & 2.2 & 78.5 & 74.9 & 82.1\\\hline
L3-8b-VA& 15.6 & 3.2 & 10.1 & 12.8 & 1.0 & 48.5 & 70.2 & 73.6\\\hline
L3-8b-FT& 28.3 & 6.9 & 14.7 & 13.7 & 1.6 & 76.5 & 76.7 & 81.2\\\hline
DPO (L3-8b)& \textbf{38.2} & \textbf{8.9} & \textbf{18.7} & \textbf{19.9} & \textbf{3.4} & \textbf{79.6} & \textbf{76.8} & \textbf{88.8}\\\hline
KTO (L3-8b)& 28.1 & 6.3 & 14.6 & 14.2 & 1.4 & 77.3 & 74.2 & 86.4\\\hline
GPT-3.5 (1-shot)& 26.1 & 6.5 & 13.6 & 14.1 & 1.3 & 75.3 & 74.8 & 78.3\\\hline
GPT-4 (1-shot)& 32.3 & 7.1 & 16.8 & 17.9 & 2.1 & 77.2 & 75.2 & 81.6\\\hline
Gemini (1-shot)& 27.3 & 6.9 & 15.1 & 16.4 & 1.6 & 76.4 & 75.8 & 80.5\\\hline
    \end{tabular}
    \end{adjustbox}
    \vspace{-2mm}
    \caption{Traditional, UniEval and GPT-4 Critique evaluations on Research Presentation Dataset, (The best outcomes are marked in \textbf{Bold}) Additional Results of Table \ref{tab:rouge_gpt}}
    \label{tab:appendix-neurips_rouge_gpt}
\end{table*}

\begin{table*}[!ht]
    \centering
    \begin{adjustbox}{width=0.4\linewidth}
    \begin{tabular}{|c|c|c|}
        \hline
        \textbf{Dataset} & \textbf{Model} &  \textbf{Rat}\\
        \hline
        
News & DPO (L2-7b) & 91.6\\\hline
News & DPO (Mst-7b) & \textbf{94.6}\\\hline
News & DPO (L3-8b) & 93.5\\\hline\hline
CSPubSum & DPO (L2-7b) & 85.3\\\hline
CSPubSum & DPO (Mst-7b) & 84.6\\\hline
CSPubSum & DPO (L3-8b) & \textbf{87.1}\\\hline\hline
Wikipedia & DPO (L2-7b) & 58.3\\\hline
Wikipedia & DPO (Mst-7b) & 69.2\\\hline
Wikipedia & DPO (L3-8b) & \textbf{71.4}\\\hline\hline
Research Presentation & DPO (L2-7b) & 83.9\\\hline
Research Presentation & DPO (Mst-7b) & 84.2\\\hline
Research Presentation & DPO (L3-8b) & \textbf{87.3}\\\hline
    \end{tabular}
    \end{adjustbox}
    \caption{Gemini Critique on DPO variants
    }
    \vspace{-1mm}
    \label{tab:gemini-dpo}
\end{table*}

\begin{table*}[!ht]
    \centering
    \begin{adjustbox}{width=0.85\linewidth}
    \begin{tabular}{|c|c|c|c|c|c|c|c|c|c|}
        \hline
        \multirow{2}{*}{\textbf{Dataset}} & \multirow{2}{*}{\textbf{Model}} & \multicolumn{6}{|c|}{\textbf{Traditional Metrics}} & \textbf{UniEval} & \textbf{GPT-4}\\
        \cline{3-9}
         &  & \textbf{R1} & \textbf{R2} & \textbf{RL}& \textbf{Mt} & \textbf{Bl} & \textbf{BeF1} & \textbf{Rel} & \textbf{Critique}\\
        \hline
\multicolumn{10}{|c|}{DPO using Vanilla Model}\\\hline
News Corpus & DPO (L3-8b)& 16.2 & 7.1 & 16.3 & 26.3 & 2.6 & 74.9 & 74.0 &  82.3\\\hline
CSPubSum & DPO (L3-8b)& 17.5 & 4.6 & 11.1 & 13.3 & 1.7 & 52.6 & 57.3 & 75.1\\\hline
Wikipedia & DPO (L3-8b)& 23.6 & 8.2 & 16.7 & 25.2 & 3.8 & 73.9 & 60.6 & 57.1\\\hline
Research Presentation & DPO (L3-8b) & 16.8 & 6.9 & 14.8 & 22.5 & 3.1 & 63.1 & 61.4 & 68.4\\\hline
\multicolumn{10}{|c|}{DPO using SFT Model on full data}\\\hline
News Corpus & DPO (L3-8b)& 18.6 & 8.6 & 19.3 & 27.3 & 2.8 & 75.1 & 75.2  & 85.2\\\hline
CSPubSum & DPO (L3-8b)& 18.4 & 6.5 & 15.1 & 22.3 & 2.7 & 56.8 & 56.6 & 76.9\\\hline
Wikipedia & DPO (L3-8b)& 23.8 & 9.3 & 18.1 & 25.7 & 4.6 & 73.1 & 59.2 & 59.3\\\hline
Research Presentation & DPO (L3-8b) & 17.5 & 7.2 & 16.3 & 24.5 & 3.7 & 64.5 & 58.2 & 69.4\\\hline
\multicolumn{10}{|c|}{DPO using 50\% data on SFT model and 50\% on DPO}\\\hline
News Corpus & DPO (L3-8b)& 36.8 & 16.0 & 28.6 & 33.5 & 6.9 & 83.6 & 84.7  & 91.3\\\hline
CSPubSum & DPO (L3-8b)& 32.3 & 9.9 & 23.0 & 24.9 & 3.5 & 82.3 & 67.8 & 89.0\\\hline
Wikipedia & DPO (L3-8b)& 42.5 & 26.3 & 38.4 & 38.7 & 16.7 & 83.4 & 67.3 & 67.3\\\hline
Research Presentation & DPO (L3-8b) &38.2 & 8.9 & 18.7 & 19.9 & 3.4 & 79.6 & 76.8 & 88.8\\\hline
    \end{tabular}
    \end{adjustbox}
    \caption{DPO variant - DPO using vanilla model, SFT using full data and SFT using 50\% of data }
    \vspace{-1mm}
    \label{tab:dpo-three-sets}
\end{table*}

\begin{table*}[!ht]
    \centering
    \begin{adjustbox}{width=0.65\linewidth}
    \begin{tabular}{|c|c|c|c|c|c|c|c|c|c|c|c|}
        \hline
        \multirow{2}{*}{\textbf{Training set}} & \multirow{2}{*}{\textbf{Test set}} &  \multicolumn{6}{|c|}{\textbf{Traditional Metrics}} &  \multicolumn{1}{|c|}{\textbf{GPT-4}}\\
        \cline{3-8}
        &  & \textbf{R1} & \textbf{R2} & \textbf{RL}& \textbf{Mt} & \textbf{Bl} & \textbf{BeF1} &  \textbf{Critique}\\
        \hline
News & CSPubSum & 28.7 & 9.4 & 21.3 & 23.2 & 3.0 & 80.2 & 71.5\\\hline
News & Wikipedia & 34.2 & 21.6 & 32.7 & 33.2 & 11.2 & 76.2 & 63.1\\\hline
CSPubSum & News & 28.3 & 9.3 & 24.2 & 27.1 & 3.4 & 79.3 & 75.0\\\hline
CSPubSum & Wikipedia & 35.1 & 19.7 & 28.9 & 29.1 & 10.3 & 75.4 & 65.8\\\hline
Wikipedia & News & 30.1 & 11.3 & 25.3 & 26.1 & 4.9 & 81.5 & 74.0\\\hline
Wikipedia & CSPubSum & 28.3 & 9.2 & 20.4 & 22.1 & 3.1 & 78.3 & 72.0\\\hline
    \end{tabular}
    \end{adjustbox}
    \caption{Cross Dataset Validation Assessment on DPO (Llama3-8b) model}
    \vspace{-1mm}
    \label{tab:out-of-domain}
\end{table*}

\section{Edge-Case Evaluation and Limitations} \label{appendix : spot_limitation}

\begin{figure*}[!ht]
    \centering
    \begin{adjustbox}{width=0.9\textwidth,center}
    \includegraphics{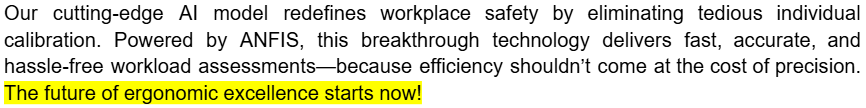}
    \end{adjustbox}
    \caption{Spotlight Example of failure case}
    \label{fig:spot_limitation}
\end{figure*}

To better understand the limitations of our approach, we conducted a qualitative analysis of failure cases and edge scenarios in model-generated spotlights. This analysis reveal some common patterns such as hallucination and misrepresentation of the source content. Fig. \ref{fig:spot_limitation} shows a spotlight which inaccurately frames the research as a commercial advertisement, reflecting a failure mode where the model prioritizes engagement over factual grounding. 

\end{document}